\documentclass[aps,prx,floatfix,reprint,superscriptaddress]{revtex4-2}
\usepackage{graphicx}%
\usepackage{multirow}%
\usepackage{amsmath,amssymb,amsfonts}%
\usepackage{amsthm}%
\usepackage{dsfont}%
\usepackage{mathrsfs}%
\usepackage{physics}%
\usepackage[title]{appendix}%
\usepackage{xcolor}%
\usepackage{booktabs}%
\usepackage{hyperref}
\usepackage{xr-hyper}
\DeclareRobustCommand{\okina}{%
	\raisebox{\dimexpr\fontcharht\font`A-\height}{%
		{`}\;%
	}%
}

\DeclareMathOperator{\diag}{diag}

\begin{document}

%\preprint{}

\title{Origins and mitigation of double descent in reduced order modeling}

% another option is "A stochastic framework for normal aortic morphodynamics across the human lifespan"



\author{Andrei A. Klishin}
\thanks{To whom correspondence should be addressed: \\
aklishin@hawaii.edu, kmanohar@uw.edu}
\affiliation{Department of Mechanical Engineering, University of Hawai\okina i at M\={a}noa, Honolulu, HI}
\affiliation{AI Institute in Dynamic Systems, University of Washington, Seattle, WA 98195, USA}
\affiliation{Department of Mechanical Engineering, University of Washington, Seattle, WA 98195, USA}

\author{J.~Nathan Kutz}
\affiliation{AI Institute in Dynamic Systems, University of Washington, Seattle, WA 98195, USA}
\affiliation{Departments of Applied Mathematics and Electrical and Computer Engineering, University of Washington, Seattle, WA 98195, USA}

\author{Krithika Manohar}
\affiliation{AI Institute in Dynamic Systems, University of Washington, Seattle, WA 98195, USA}
\affiliation{Department of Mechanical Engineering, University of Washington, Seattle, WA 98195, USA}


\date{\today}

\begin{abstract}
Latent low-dimensional structure in datasets of natural and engineered systems enables their sparse sensing, or full-state reconstruction from historical data and very few carefully chosen localized measurements. Depending on the reconstruction algorithm, sensor locations, and measurement noise, the reconstruction risk curves demonstrate a diversity of patterns including a dramatic peak in error known as double descent in Machine Learning literature. Here we explore those scenarios under a unified Data-Noise Averaging theory. Qualitatively, we formulate sufficient criteria for double descent to emerge through a catastrophic amplification of a pathological signal in reconstruction. Quantitatively, we predict the detailed risk curves at a fraction of computational cost, trace reconstruction instability to individual sensors and their combinations, and provide regularization mechanisms to mitigate the instability. We demonstrate results for both static reconstruction of Sea Surface Temperature patterns and time integration of a reduced order model of a PDE.
\end{abstract}


\maketitle

\section{Introduction}

Reduced Order Models (ROMs) are one of the centerpiece techniques of modern scientific computing as in many cases they can drastically reduce the compute and memory requirements of a simulation with only a modest or negligible reduction in accuracy \cite{lucia2004reduced, yu2019ROM, kutz2023machine}. A key property of ROMs is \emph{sparsity}, which comes in two flavors. On one side, the entire high-dimensional $n\gg 1$ state of a system can be efficiently approximated with just a few linear modes or nonlinear latent parameters. On the other side, in order to avoid evaluating expensive functions many times, ROMs perform computations in only a sparse set of spatial locations and interpolate the remaining state space. Mode-sparsity and space-sparsity compete and intersect in a complex way across both sparse sensing of static snapshots \cite{manohar2018data} and more general ROMs of dynamical systems \cite{barrault2004empirical, chaturantabut2009discrete}.

% modes or latent dimensions
% sampling points

% empirical approximation
% While on one side the full state of the system, or the nonlinear term of a model

% why reduced order modeling
% can reconstruct the full state from sparse local measurements, can also approximate nonlinear terms in dynamics

The same problem of approximately reconstructing the state from sparse observations is known as \emph{empirical interpolation}~\cite{barrault2004empirical, chaturantabut2009discrete, drmac2016qdeim} or \emph{state estimation}~\cite{Kakasenko2026BridgingTG} in applied and computational mathematics literature  and as \emph{sensor placement} in engineering \cite{manohar2018data, karnik2024energies}; for consistency, we use the latter terminology in this paper. Early work on empirical interpolation required the number of sensors $p$ to be larger than the number of reconstructed modes $r$ in order for reconstruction matrices to be properly invertible with $p\geq r$ (though some methods use sparse time series to reconstruct even more modes \cite{farazmand2024sparse, farazmand2026state}). From this requirement many successive methods were proposed for picking the best points for evaluation, either randomly \cite{everson1995karhunen}, based on local maxima of the modes \cite{yildirim2009efficient}, with greedy condition number optimization \cite{willcox2006unsteady, chaturantabut2009discrete}, QR pivoting of a mode matrix \cite{drmac2016qdeim, manohar2018data}, or interaction landscapes \cite{klishin2023data}. Once one of these algorithms returns an ordered set of sampling locations, the Root Mean Square Error (RMSE) can be evaluated by reconstructing states from a withheld testing data set \cite{loog2019minimizers, viering2022shape}. Reconstruction is trivially extended to the undersampled case $p<r$ through the Moore-Penrose pseudoinverse, resulting in a continuous \emph{risk curve} of RMSE against $p$ at fixed $r$. Empirical and benchmarking results show the curve having a sharp cusp near $p\approx r$, the boundary of undersampled and oversampled regime \cite{clark2020multi}, for instance in the reference documentation of a popular sparse sensing package \texttt{PySensors} \cite{desilva2021pysensors,pysensors2020ddes}, though it was not discussed as a distinct phenomenon in the literature.

% risk curve
% though extension to $p<r$ is easy with regularization or pseudoinverse

% Computationally, extension of reconstruction to the undersampled regime $p<r$ is trivial to 
% Empirical results showed sharp cusps at $p\approx r$
% The cusp was not discussed as a distinct phenomenon

The cusp or singularity of the risk curve frequently appears in supervised machine learning, popularized under the name ``double descent'' since the curve initially falls, then abruptly spikes, and falls a second time \cite{belkin2019reconciling}. While the current wave of research has been primarily driven by the neural network applications deep into the overparameterized regime \cite{belkin2019reconciling, nakkiran2021deep}, the phenomenon itself has been observed in linear regression and perceptrons since at least the 1980s \cite{breiman1983many, vallet1989linear, opper1990ability, watkin1993statistical, raudys1998expected, duin2000classifiers, loog2020brief}. In exotic cases the risk curve can show triple \cite{dascoli2020triple} or even multiple descent \cite{transtrum2025generalized}. Qualitatively, the cusp is explained by the model overfitting the training set when the number of model parameters exactly matches the number of data points (known as the interpolation threshold) \cite{belkin2019reconciling}, and can be mitigated by parameter regularization \cite{nakkiran2021optimal}, early stopping of the training \cite{nakkiran2021deep}, or addition of unlabeled data points in semi-supervised learning \cite{krijthe2016peaking}. Quantitatively, under weak feature assumptions the cusp shape can be predicted by bias-variance decomposition and averaging over the random data matrices \cite{dascoli2020double, belkin2020two, hastie2022surprises, mei2022generalization}, with connections to statistical physics and jamming phenomena \cite{spigler2019jamming, rocks2022memorizing}. Cusp locations can also be predicted by the bounds set by the Generalized Aliasing Decomposition (GAD) \cite{transtrum2025generalized}.
% the shape of the curve for weak features is expressed by the bias

% based on interpolation and oversampling. lot of attention to how to pick the proper points (DEIM, QDEIM, QR sensors, landscapes)
% machine learning has renewed its interest in the risk curves in generalization (RMSE on the withheld data set), notably the peak when number of parameters matches the number of training data points, visible when either of the parameters is held fixed while the other one varies, in both linear regression and nonlinear neural networks. empirically the risk curve initially drops as the number of data points increases from zero, then shows a sharp peak at p~r, and falls again, hence it is popularly known as "double descent". while the current wave of research has been primarily driven by the neural network applications, the phenomenon itself has been observed in linear regression and perceptrons since at least the 1980s (loog and early papers).
% in exotic cases can have triple (nakkiran) or multiple descent (dascoli). explained by overfitting to the training set in supervised learning at the interpolation threshold (belkin), mitigated by regularization (nakkiran) or early stopping of the training.

The vast majority of double descent studies focus on supervised learning, i.e. using features to predict labels, and observe a cusp either model-wise (increasing number of parameters), sample-wise (increasing amount of data), or epoch-wise (increasing training length) \cite{belkin2019reconciling, rahimi2024unveiling}. Rare studies consider double descent in unsupervised learning, such as discovery of latent low-dimensional structure in the data via an autoencoder \cite{rahimi2024unveiling}. The ROM case combines elements of both, as it first finds the latent structure (such as a modal decomposition), and then selects a few of the features (sensor locations) to predict the rest of the features (different from principal component regression \cite{Gedon2024PCA}). In addition, many machine learning studies of double descent rely on weak features that achieve prediction in the aggregate and can be chosen essentially randomly; in contrast, in sparse sensing, the sensors are chosen via a specific algorithm, akin to the ``prescient features'' \cite{breiman1983many, belkin2020two}.

In this paper, we compute the risk curve of sparse sensing via a covariance decomposition of the reconstruction error, and use this decomposition to explain the origins of double descent and its mitigation. Our main contribution is the Data-Noise Averaging (DNA) theory of reconstruction risk, organized as follows:
\begin{enumerate}
    \item \textbf{Closed-form analytical risk prediction.} We construct an explicit analytic expression for the reconstruction risk curve that exploits the data singular values, rank $r$, and sensor selection matrix directly, without averaging over data matrices drawn from a generic ensemble. The expression introduces no free parameters beyond the design choices of the sparse sensing setup, and predicts both the location and magnitude of the double descent spike across all tested configurations.
    \item \textbf{Computational efficiency.} The DNA theory reduces risk curve computation to efficient low-rank linear algebra operations, achieving a $10^3\times$ reduction in cost relative to empirical risk averaging, reducing computation from approximately 50 minutes to 3 seconds on a standard laptop for Sea Surface Temperature reconstruction ($n=44219,r=100$).
    \item \textbf{Reconstruction instability.} For either random or intelligent sensor placement, we show that the contamination from truncated modes and measurement noise is resonantly amplified when the inversion matrix $\mathbf{M}$ develops low-lying eigenvalues, leading to a sharp instability peak. We diagnose two distinct failure cases using the Gram matrix of sensing vectors: \textit{1-point failure} driven by individual sensors, and \textit{ensemble failure} driven by collective sensor interactions. The sampling regime $p=r$ common in ROMs \cite{chaturantabut2009discrete, drmac2016qdeim, manohar2018data} is identified as the worst possible operating point, arising from an inevitable orthogonality crisis in $\mathbb{R}^r$.
    \item \textbf{Mitigation and design.} The shape of the risk curve also suggests remedies, such as surprisingly using \emph{fewer} sensors. We demonstrate that a combination of undersampling with regularized optimal linear estimation achieves lower reconstruction error and reduced computational cost in ROMs.
\end{enumerate}

% Unlike the principal component regression \cite{Gedon2024PCA}, the prediction ``labels'' are not truly distinct from the ``features''. 
% here we compute the risk curve by decomposing it into different terms (dascoli). instead of averaging over the data matrices drawn from an ensemble, we use a specific benchmark dataset so that specific chosen points can be blamed. we show that precisely at the interpolation threshold the contamination from unmodeled modes and measurement noise gets resonantly amplified, leading to a sharp peak. the peak is driven by the exhaustion of the sensing vectors. the common p=r regime is the worst possible choice. the shape of the curve suggests the remedies, such as surprisingly using \emph{fewer} sensing locations. we demonstrate that a combination of undersampling with regularized optimal linear estimation leads to both lower error and computational resources in reduced order modeling

The rest of the paper is organized as follows. In Sec.~\ref{sec:expt} we introduce the principal ideas behind state reconstruction from sparse measurements, formulate the risk curve shape as the central object of study, and conduct empirical tests to explain \emph{qualitatively} the reasons for the double descent phenomenon. In Sec.~\ref{sec:theory} we construct the DNA theory by computing the covariance matrix of the reconstruction error and predicting the risk curve \emph{quantitatively} through decomposition into terms, and suggest mitigation mechanisms. In Sec.~\ref{sec:sparsesensing} we apply our theory to Case Study 1: Sparse Sensing of static snapshots and trace the effect of reconstruction design choices on the risk curves. In Sec.~\ref{sec:DEIM} we apply the reconstruction lessons to Case study 2: empirical interpolation for partial differential equations (PDEs). Sec.~\ref{sec:disc} summarizes the results and suggests future steps.

\section{Double descent in sparse sensing}
\begin{figure}
    \centering
    \includegraphics[width=\linewidth]{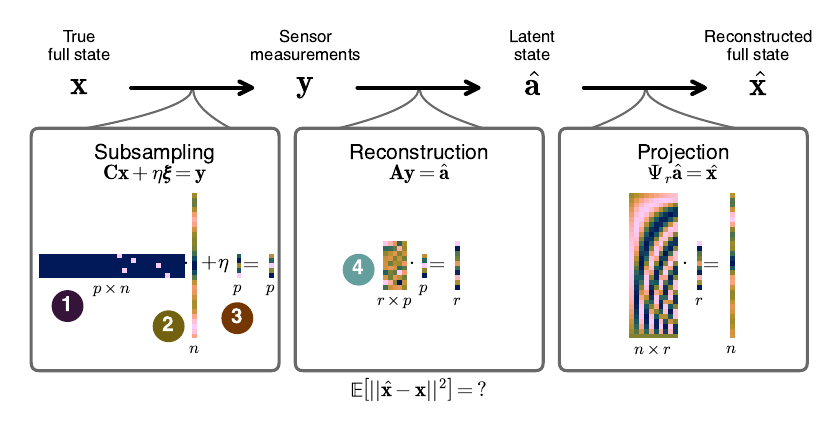}
    \caption{Computational flow of sparse sensing for reconstruction. The circled numbers highlight the junctions of qualitative system setup choices (design factors). A full state is subsampled at a small set of locations (1), sampling from either the full state or its projection on the leading modes (2), with possible measurement noise (3). The sensor measurements are converted into the latent state estimation with or without regularization (4). The latent state is then used to project back to full state space. The central question of the paper is how the design choices qualitatively and quantitatively impact the expected reconstruction error of the full state (risk curve).}
    \label{fig:diagram}
\end{figure}

\label{sec:expt}
In this section we introduce the basic scheme of state reconstruction from sparse sensors, demonstrate several different behaviors of the risk curve on an empirical dataset, and explain the double descent phenomenon \emph{qualitatively}, with a more detailed calculation following.

\subsection{Sparse sensing}
Sparse sensing is a family of methods that differ by subtle design choices. Fig.~\ref{fig:diagram} highlights those qualitative system setup choices (design factors) with the markers 1--4, explained in more detail below. The idea of sparse sensing is predicated on representing a high-dimensional system state $\mathbf{x}$ as a sparse combination of a small number of modes~\cite{manohar2018data}. The modes are usually learned through Proper Orthogonal Decomposition (POD) of a training data set of state snapshots $\mathbf{X}_{train}:n\times N$, where $n$ is the state dimension, $N$ is the number of snapshots, and $n>N$ for most applications. The POD allows us to write any state as a linear combination of the modes:
\begin{align}
    \mathbf{X}_{train}=\mathbf{U}\boldsymbol{\Sigma} \mathbf{V}^\top\quad\Rightarrow\quad \mathbf{x}=\boldsymbol{\Psi}_r \mathbf{a}_r+\boldsymbol{\Psi}_c \mathbf{a}_c,
    \label{eqn:POD}
\end{align}
where we separated the left singular vector matrix $\mathbf{U}=\boldsymbol{\Psi}_r|\boldsymbol{\Psi}_c$ into the leading $r\ll N$ modes that carry the majority of signal variance, and the remaining $N-r$ modes \cite{gavish2014optimal, udell2019big, quinn2022information}. Similarly, we separate the square diagonal matrix of singular values $\boldsymbol{\Sigma}$ into the leading $r$ singular values $\boldsymbol{\Sigma}_r$ and the rest $\boldsymbol{\Sigma}_c$. We further normalize the singular values $\boldsymbol{\Sigma}'_r= \boldsymbol{\Sigma}_r/\sqrt{N-1}$, $\boldsymbol{\Sigma}'_c= \boldsymbol{\Sigma}_c/\sqrt{N-1}$ following the prescription in Ref.~\cite{Kakasenko2026BridgingTG}: while the original singular values express the covariance structure of the \emph{whole data matrix}, the normalized singular values express the covariance of \emph{typical states}. We note that following the initial POD and normalization, the number of training snapshots $N$ is not used again, thus making it irrelevant to reconstruction instabilities.

While the full training data matrix is available for offline calculation of the singular vectors, for the states we only have access to sparse and noisy local observations:
\begin{align}
    \mathbf{y}=\mathbf{C}\mathbf{x}+\Delta\mathbf{y},
    \label{eqn:sampling}
\end{align}
where $\mathbf{C}:p\times n$ is the selection matrix consisting of $p$ rows of the identity matrix corresponding to placed sensors (factor 1), $\mathbf{x}:n$ is the full-state vector (factor 2), and $\Delta \mathbf{y}$ is measurement noise (factor 3), which is assumed to be Gaussian and uncorrelated with variance $\eta^2$.

To get a full state estimate from the sparse observations, we compute a linear estimator of the latent vector of leading mode coefficients and project it to the leading modes:
\begin{align}
    \hat{\mathbf{a}}_r=\mathbf{A}\mathbf{y};\quad \hat{\mathbf{x}}=\Psi_r \hat{\mathbf{a}}_r,
    \label{eqn:reconstruction}
\end{align}
where $\mathbf{A}:r\times p$ is a reconstruction matrix that depends on the sensor placement $\mathbf{C}$. There are different options for the reconstruction matrix $\mathbf{A}$ (factor 4), but they are all assumed to be full rank $rank(\mathbf{A})=\min(r,p)$.

The quality of reconstruction can be measured by the Root Mean Squared Error, averaged across either the training set or the holdout test set and normalized by the state dimension:
\begin{align}
    Rel.RMSE\equiv \frac{1}{\sigma_{scale}}\sqrt{\frac{1}{n}\mathbb{E}_{\mathbf{x}_{test}} [\norm{\hat{\mathbf{x}}-\mathbf{x}}^2] }
    \label{eqn:RMSE}
\end{align}
where $\sigma_{scale}$ is the natural scale of the dataset given by the standard deviation of the training set $\mathbf{X}_{train}$. The benefit of normalization lies in that the most naive reconstruction guess $\hat{\mathbf{x}}=0$ would result in expected $Rel.RMSE=1$, which makes it easy to judge the relative success of different methods. Relative RMSE can be computed for a sequence of sensor sets and form a \emph{risk curve} with sensor number $p$ serving as the horizontal axis, which can have a variety of qualitatively different shapes.

\subsection{Binary design factors}
\begin{figure*}
    \centering
    \includegraphics[width=.8\linewidth]{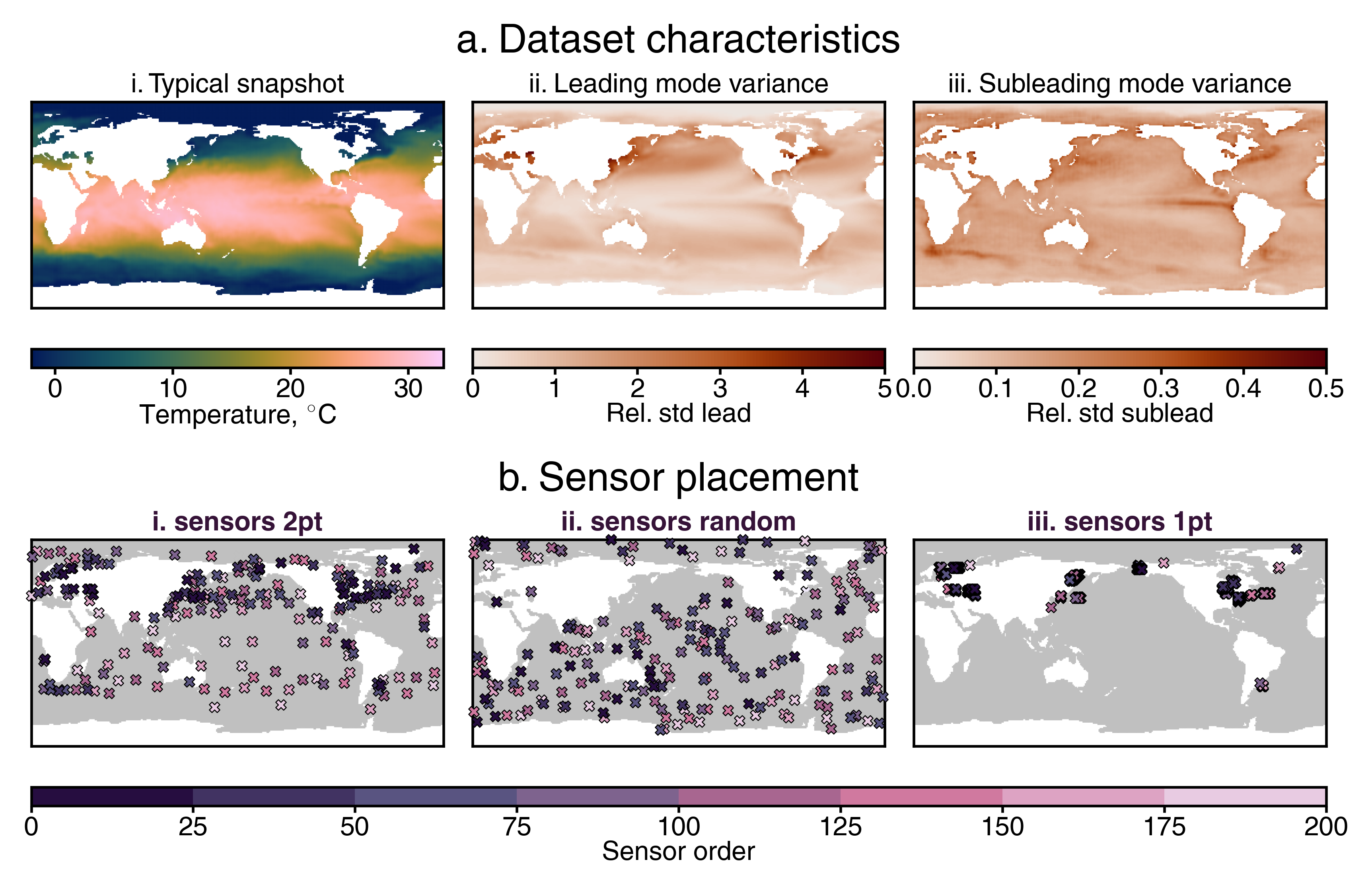}
    \caption{Sea Surface Temperature dataset and sparse sensors placement. (a) Dataset properties: (a.i) a typical snapshot of raw temperature distribution in $^\circ \text{C}$, (a.ii) the standard deviation of data in each pixel within the leading $r=100$ modes, (a.iii) the standard deviation of data within the subleading modes beyond the top 100 (contamination). (b) Sensor placement patterns under different methods: (b.i) 2-point, (b.ii) random, (b.iii) 1-point. Sensor are color coded by placement order in batches of 25.}
    \label{fig:SST}
\end{figure*}
The above derivation sketch deliberately omitted several experimental design choices. In order to gain a \emph{qualitative} intuition for the shape of the risk curve, we consider the experimental design space to consist of four binary factors:
\begin{enumerate}
    \item \textbf{Sensor placement algorithm}. Previous studies highlighted the critical importance of choosing the sensor locations given a tailored sparse basis, with a variety of available algorithms \cite{everson1995karhunen, willcox2006unsteady, yildirim2009efficient, chaturantabut2009discrete, drmac2016qdeim, manohar2018data}. First, we use the 2-point algorithm of Ref.~\cite{klishin2023data}, closely similar to QR pivoting \cite{manohar2018data} and implemented in \texttt{PySensors 2} package \cite{karnik2025pysensors2}, which gives a sequence of locations aiming to maximize the captured signal variance while minimizing sensor crosstalk. Second, we use a single realization of sensors drawn at random without replacement. Both algorithms return $p_{max}$ sensors in a priority order, so that computations can be carried out for the first $p$ sensors as $p$ gradually increases.
    
    \item \textbf{State projection}. We separate the original data matrix $\mathbf{X}$ into the training and test matrices $\mathbf{X}_{train}$, $\mathbf{X}_{test}$ randomly in 80:20 proportion. The training data matrix is used to compute the POD and thus the leading singular vectors $\boldsymbol{\Psi}_r$. We compute RMSE on both training and test data sets. First, we consider each state (matrix column) as is, and second we project the data matrices onto the leading $r$ modes $\mathbf{X}\to\boldsymbol{\Psi}_r(\boldsymbol{\Psi}_r \mathbf{X})$ to limit the spectral content of the signal to exactly $r$ modes. The bracket enforces multiplication order to avoid an intermediate rank $r$ matrix of size $n\times n$.

    \item \textbf{Amount of noise}. First, we use Gaussian uncorrelated noise with the covariance matrix $\expval{\Delta \mathbf{y}\Delta \mathbf{y}^\top}=\eta^2 \mathbb{I}_p$ and set the noise level equal to half of data standard deviation $\eta_{noise}=0.5 \sigma_{scale}$. Second, we use zero noise.
    
    \item \textbf{Linear estimator}. We consider two different choice of the linear estimator matrix $\mathbf{A}:r\times p$. In both cases we assemble the sensing matrix $\boldsymbol{\Theta}\equiv \mathbf{C}\boldsymbol{\Psi}_r$, which consist of subsampled \emph{rows} of the leading singular vector matrix. First, we use the least squares estimator via the Moore-Penrose pseudoinverse $\mathbf{A}_L=\boldsymbol{\Theta}^\dagger$. Second, we use the Bayesian optimal linear estimator that includes the prior information of state variance $\mathbf{A}_R=(\eta_{reg}^2 {\boldsymbol{\Sigma}'_r}^{-2}+\boldsymbol{\Theta}^\top \boldsymbol{\Theta})^{-1} \boldsymbol{\Theta}^\top$, where $\boldsymbol{\Sigma}'_r$ are the normalized leading singular values of the training set and $\eta_{reg}$ is the expected noise magnitude (derived in Ref.~\cite{klishin2023data} with normalization from Ref.~\cite{Kakasenko2026BridgingTG}). The optimal estimator is essentially a ridge (Tikhonov) regularized regression where the expected noise magnitude regulates the strength of regularization.
    
\end{enumerate}

We test all $2^4=16$ combinations of the binary factors on the Sea Surface Temperature data set by NOAA \cite{huang2021improvements} that consists of 1727 weekly snapshots between Dec 31, 1989 and Jan 1, 2023 that are separated into $N_{train}=1381$ train and $N_{test}=346$ test states. Each snapshot contains the interpolated water temperature in degrees Celsius with spatial resolution of $1\times 1$ degree (longitude by latitude, Fig.~\ref{fig:SST}a). After filtering out the part of Earth covered by ground this leaves the dimension $n=44219$ for each state snapshots. The standard deviation of the dataset is $\sigma_{scale}=1.945^\circ \text{C}$. This dataset demonstrates a steep decay of singular values so we choose $r=100$ states for the reduced order model. We use three methods to place $p_{max}=200$ sensors (Fig.~\ref{fig:SST}b): the 2-point method starts with placing sensors on continental coasts and in inland bodies of water and then gradually fills out the open ocean; the random method drops sensors in random order; the 1-point method (known to be strictly inferior \cite{klishin2023data}) ignores sensor correlations and places sensors in tight clusters in locations of maximal signal variance.

\begin{figure*}
    \centering
    \includegraphics[width=.8\linewidth]{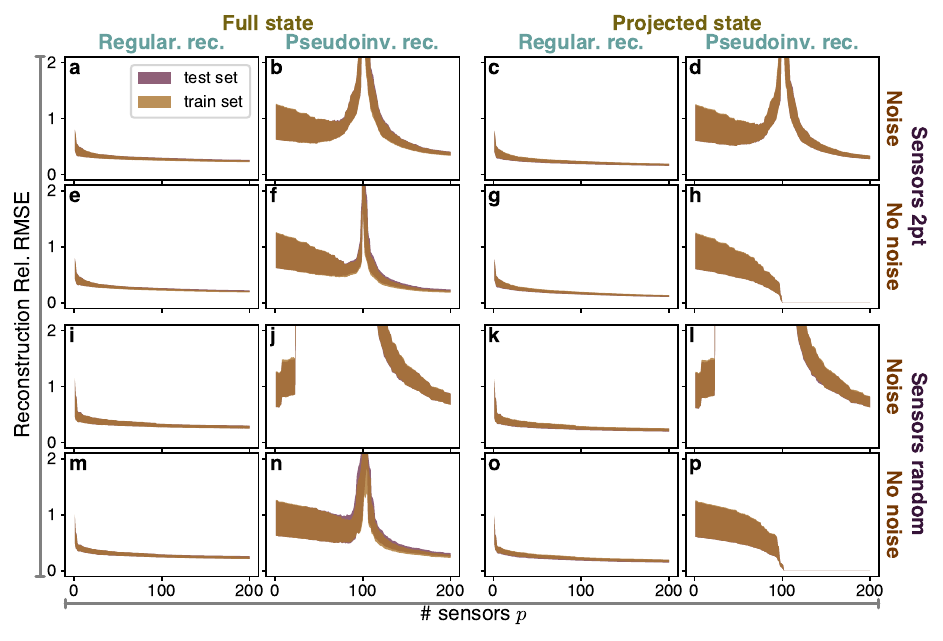}
    \caption{Qualitative enumeration of the risk curves of sparse sensing. Each design parameter of Fig.~\ref{fig:diagram} takes one of two values, resulting in $2^4=16$ combinations of all possible plots. The design choices are indicated on the labels of panel rows and columns. Each panel uses $r=100$ modes for reconstruction on the SST dataset and up to $p_{max}=200$ sensors.
    Purple and brown show the empirical mean RMSE$\pm$one standard deviation, across test and train sets, respectively. The two shaded areas overlap almost everywhere in the figure.}
    \label{fig:factorial}
\end{figure*}

The results of the enumeration are presented in Fig.~\ref{fig:factorial} where the risk curve either decreases monotonically or first decreases, then abruptly spikes, and decreases again in what we term \emph{double descent}. The pattern of risk curves suggests a qualitative explanation of the origins of double descent. Using regularization completely suppresses the spike. If regularization is absent, the risk curve can be interpreted as a ``pathological signal'' entering an ``amplifier''. Here the pathological signal is anything not in the modeled $r$ modes: either the subleading modes, or the measurement noise, or both. The amplifier somehow results from the collection of sensors either close to the number of modes $p\approx r$ (panels b,d,f,n) or sometimes earlier (panels j,l). When the pathological signal and the amplifier are both present, the small errors in state measurement are disproportionally magnified, resulting in the instability. If there is no pathological signal, using $p\geq r$ reconstructs the signal perfectly ($Rel.RMSE=0$ in panels n,p).

In the following section we seek to quantify this intuition and get a handle on the functional shape of these empirical curves.

% panels jl early spike
% fully explained hp

% monotonically decreases or demonstrates a spike. The qualitative expl

% \aak{insert the figure with the empirical curves}

% observations:
% peak happens when either noise or contamination are present
% blow up is much earlier on random sensors
% regularization fully removes the peak

% SVD and reduced order representation, notation
% modes count, separation of singular values, noisy measurements
% subsampling via sensors

% different linear reconstructors, full rank

% computing RMSE per pixel for motivation

% four factors: sensors, estimators, noise, projection
% for qualitative intuition, quantitative argument below, predicting the risk curve

% SST dataset

\section{DNA theory of double descent}
\label{sec:theory}

Here we provide a \emph{quantitative} theory of double descent and its suppression. The derivation relies on decomposing the covariance matrix of the state reconstruction error into distinct terms referring to \textbf{D}ata distribution and \textbf{N}oise realizations, and then \textbf{A}veraging (hence the DNA name). Our computation predicts the overall risk curve for any sensor number and configuration. We emphasize that the derivation introduces \emph{no free parameters} beyond the design factors of the sparse sensing setup.

Crucially, extant theories of double descent rely heavily on the random matrix theory and random sampling of features \cite{dascoli2020double, belkin2020two, hastie2022surprises, mei2022generalization, rocks2022memorizing}. While such analyses can predict quantitatively the shape of the risk curve, they necessarily suppress the role of \emph{specific} sensor configurations that are relevant in applied data scenarios. Our computation expresses the covariance matrix directly in terms of the data set POD and sensor selection matrices, and thus allow tracing the state reconstruction instability back to the design decisions of the reconstruction algorithm. GAD theory similarly traced the risk peaks to the structure of the selected features and used matrix norms to accurately predict the \emph{locations} of all risk peaks \cite{transtrum2025generalized}; in contrast, the DNA theory aims to quantitatively predict the \emph{exact shape of the risk curve}.

\subsection{Error covariance matrix}

We seek to compute the error covariance matrix in the following form:
\begin{align}
    \mathbf{K}=\mathbb{E}_{\mathbf{x},\Delta \mathbf{y}} [(\hat{\mathbf{x}}-\mathbf{x})(\hat{\mathbf{x}}-\mathbf{x})^\top],
\end{align}
where the state $\mathbf{x}$ and the noise realization $\Delta \mathbf{y}$ are uncorrelated from each other. In order to compute the average, we need to directly relate the estimator $\hat{\mathbf{x}}$ to the state $\mathbf{x}$, as well as make assumptions on the covariance matrix of $\mathbf{x}$ itself.

The estimator can be expressed by plugging Eqns.~\ref{eqn:POD} and \ref{eqn:sampling} into Eqn.~\ref{eqn:reconstruction}:
\begin{align}
    \hat{\mathbf{x}}=\boldsymbol{\Psi}_r \mathbf{A}(\mathbf{C}\boldsymbol{\Psi}_r \mathbf{a}_r+\mathbf{C}\boldsymbol{\Psi}_c \mathbf{a}_c + \Delta \mathbf{y}).
\end{align}

The state covariance matrix can be estimated by the sample covariance matrix of the training data set, normalized by the number of training snapshots \cite{Kakasenko2026BridgingTG}:
\begin{align}
    &\mathbb{E}_\mathbf{x} [\mathbf{x}\mathbf{x}^\top]\approx \tfrac{1}{N-1} \mathbf{X}_{train} \mathbf{X}_{train}^\top=\tfrac{1}{N-1} \mathbf{U}\boldsymbol{\Sigma}^2 \mathbf{U}^\top  \nonumber\\ 
    &=\boldsymbol{\Psi}_r {\boldsymbol{\Sigma}'_r}^2 \boldsymbol{\Psi}_r^\top +  {\boldsymbol{\Psi}_c \boldsymbol{\Sigma}'_c}^2 \boldsymbol{\Psi}_c^\top \label{eqn:samplecov}\\
    &\mathbb{E}_\mathbf{x} [\mathbf{a}_r\mathbf{a}_r^\top]={\boldsymbol{\Sigma}'_r}^2;\; \mathbb{E}_\mathbf{x} [\mathbf{a}_c\mathbf{a}_c^\top]={\boldsymbol{\Sigma}'_c}^2;\; \boldsymbol{\Psi}_r \boldsymbol{\Psi}_c^\top=0;\label{eqn:acov} \\
    &\mathbb{E}_{\mathbf{x},\Delta y} [\mathbf{a}_c\Delta \mathbf{y}^\top]=0;\;
    \mathbb{E}_{\mathbf{x},\Delta y} [\mathbf{a}_r\Delta \mathbf{y}^\top]=0;\nonumber \\
    &\mathbb{E}_{\Delta y} [\Delta \mathbf{y}\Delta \mathbf{y}^\top]=\eta^2 \mathbb{I}_p,\label{eqn:dycov}
\end{align}
where the second line expresses the diagonal covariances of the reduced order coefficients as well as ensures that the low and high singular vectors are mutually orthogonal, while the third line expresses the covariance between sampled states and noise.

The data and noise covariances allow a compact form of writing the reconstruction covariance as a sum of six terms:
\begin{align}
    \mathbf{K}=&\underbrace{\boldsymbol{\Psi}_c \mathbf{B}_0 \mathbf{B}_0^\top \boldsymbol{\Psi}_c^\top}_\text{subleading modes} + \underbrace{\boldsymbol{\Psi}_r \mathbf{B}_1 \mathbf{B}_1^\top \boldsymbol{\Psi}_r^\top}_\text{leading modes}  \nonumber \\
    +& \underbrace{\boldsymbol{\Psi}_r \mathbf{B}_2 \mathbf{B}_2^\top \boldsymbol{\Psi}_r^\top}_\text{contamination} + \underbrace{\boldsymbol{\Psi}_r \mathbf{B}_3 \mathbf{B}_3^\top \boldsymbol{\Psi}_r^\top}_\text{noise} \nonumber \\
    -&\underbrace{\boldsymbol{\Psi}_c \mathbf{B}_0 \mathbf{B}_2^\top \boldsymbol{\Psi}_r^\top - \boldsymbol{\Psi}_r \mathbf{B}_2 \mathbf{B}_0^\top \boldsymbol{\Psi}_c^\top}_\text{cross terms}
    \label{eqn:K}\\
    \mathbf{B}_0=&\boldsymbol{\Sigma}'_c;\quad \mathbf{B}_1=(\mathbf{A}\boldsymbol{\Theta}-\mathbb{I}_r)\boldsymbol{\Sigma}'_r;\label{eqn:B01}\\
    \mathbf{B}_2=& \mathbf{AC} \boldsymbol{\Psi}_c \boldsymbol{\Sigma}'_c;\quad \mathbf{B}_3=\eta  \mathbf{A},\label{eqn:B23}
\end{align}
where the $\mathbf{B}_0$ term contains the subleading modes of the reconstructed signal that cannot be expressed by the first $r$ singular vectors; the $\mathbf{B}_1$ term quantifies how well the true coefficients within the leading $r$ modes are reconstructed; the $\mathbf{B}_2$ term expresses the contamination of the measured signal by the subleading modes; the $\mathbf{B}_3$ term expresses the contamination of the measured signal by the sensor noise; the cross terms between $\mathbf{B}_{0,2}$ affect the covariance but not the overall error as explained below. We provide a detailed interpretation of each term's effect in the following subsections and demonstrate that the double descent phenomenon is mostly driven by the $\mathbf{B}_2$ and $\mathbf{B}_3$ terms similarly to the ``double trouble'' argument of Ref.~\cite{dascoli2020double}. The terms also neatly correspond to those in the GAD theory \cite{transtrum2025generalized}: model insufficiency is the part of the signal that cannot be expressed by the model (our $\mathbf{B}_0$), data insufficiency is the degree to which parameters are under-constrained by too little data (our $\mathbf{B}_1$), and generalized aliasing expresses the ``crosstalk'' between modeled and unmodeled parts of the signal (our $\mathbf{B}_{2,3}$). Before detailed interpretation of each term , we show how the expression $\mathbf{K}$ can be used for analytic predictions.

Note that the matrix $\mathbf{K}$ is of size $n\times n$ but of rank no more than $N$, depending on the rate of decay of the training data singular values. Holding the whole covariance matrix in computer memory is thus inefficient, but it can be used to compute summary expressions. In particular, the prediction of average reconstruction error is given by:
\begin{align}
    \widehat{Rel.RMSE}=\frac{1}{\sigma_{scale}}\sqrt{\frac{1}{n}\Tr \mathbf{K}},
\end{align}
where we can use the fact that $\boldsymbol{\Psi}_r$ and $\boldsymbol{\Psi}_c$ are isometries and do not change the 2-norm of the vector they act upon. In this case the RMSE prediction takes a particularly simple form:
\begin{align}
\boxed{\widehat{RMSE}=
    \sqrt{\frac{1}{n}\sum_{l=0}^{3}\left( \sum_{ij}\mathbf{B}^2_{l,ij} \right)}},
    \label{eqn:RMSE_DNA}
\end{align}
% \begin{align}
% \boxed{\widehat{RMSE}=
%     \sqrt{\frac{1}{n}\sum_{l=0}^{3}\left( \sum_{ij}\mathbf{B}^2_{0,ij} + \sum_{ij}\mathbf{B}^2_{1,ij} + \sum_{ij}\mathbf{B}^2_{2,ij} +\sum_{ij}\mathbf{B}^2_{3,ij} \right)}},
%     \label{eqn:RMSE_DNA}
% \end{align}
where the squaring is applied to the $\mathbf{B}$ matrices elementwise; the $\mathbf{B}_l$ matrices are of different sizes, so the $ij$ summation needs to be performed first within each matrix. Note that the cross terms are traceless because of orthogonality $\boldsymbol{\Psi}_r \boldsymbol{\Psi}_c^\top$ and do not contribute to the RMSE prediction. The expression \eqref{eqn:RMSE_DNA} consists of efficient low-rank linear algebra operations that can be computed for any choice of sensor and linear estimator matrix $\mathbf{A}$, and thus can be used for exploration of the qualitative behaviors of the risk curve.

\subsection{Computational complexity}
\paragraph{Assumptions}
In order to assess the relative computational complexity of the DNA theory as opposed to naive risk curve averaging, we introduce several accounting assumptions: (i) only multiplications are counted, ignoring scalar-by-scalar $\order{1}$ operations, (ii) row-subsampling a matrix (i.e. left multiplications by $\mathbf{C}$) is negligibly cheap, (iii) dense matrix multiplication is cubic in matrix dimensions, multiplying by a diagonal matrix is quadratic, (iv) an $a\times b$ matrix can be (pseudo-)inverted in $\order{ab\cdot \min(a,b)}$, (v) We use $r_c\leq N-r$ subleading singular values, the number of which can be reduced if singular values are truncated at machine precision level. Under these assumptions, the cost of obtaining an error estimate at a given sensor budget $p$ can be computed and added up for $p\in[1,p_{max}]$ to obtain the full risk curve, assuming $p_{max}>r$ to investigate the oversampled regime.

\paragraph{Reconstruction operator}
The pseudoinverse reconstruction $\mathbf{A}_L:r\times p$ is computed in a single operation $\order{pr\cdot \min(p,r)}$. The regularized reconstruction $\mathbf{A}_R$ requires two multiplications and inversion $\order{rpr+r^3+r^2p}$ (assuming that adding the diagonal term $\eta^2_{reg}{\boldsymbol{\Sigma}'}^{-2}_r$ is negligible in comparison). Since the reconstruction operators need to be constructed for both naive averaging and DNA, the cost is the same at $\order{r^4+(p_{max}^2-r^2)r^2}$ for pseudoinverse and $\order{r^2 p_{max}^2+r^3 p_{max}}$ for regularized reconstruction. If the maximal oversampling is not very large, e.g. $p_{max}\approx 2r$, the costs of both reconstruction approaches are equivalent.

\paragraph{Naive risk curve}
For the naive reconstruction at fixed $p$ the reconstruction $\hat{\mathbf{X}}=\boldsymbol{\Psi}_r \mathbf{ACX}$ takes $\order{nrpN}$ operations where $N$ is the test set size. Adding measurement noise to the sensed locations takes another $\order{pN}$ and evaluating the RMSE between the true and reconstructed states takes another $\order{nN}$ multiplications. Aggregating over all $p$ values to form a risk curve, the total cost of naive averaging is $\order{nrp_{max}^2 N+p_{max}^2 N+np_{max}N}$.

\paragraph{DNA theory}
For the DNA theory at fixed $p$, the cost consists of evaluating all $\mathbf{B}$ matrices and squaring them elementwise. For $\mathbf{B_0}$, the cost of squaring the diagonal elements is $\order{r_c}$ and does not depend on the sensor choice, thus the result can be computed once and reused across the whole risk curve. For $\mathbf{B_1}$, the cost is $\order{rpr}$ for matrix multiplication and $r^2$ for squaring. For $\mathbf{B}_2$, the cost is $\order{rpr_c}$ for matrix multiplication and $rr_c$ for squaring. For $\mathbf{B}_3$, there is only a matrix-by-scalar multiplication and squaring for a total of $\order{rp}$. Adding up the costs of each matrix and aggregating over all $p$ values, the total cost of DNA theory is $\mathcal{O}(r_c+r^2 p_{max}^2+r^3 p_{max}+r^2 p_{max}+rr_cp_{max}^2\allowbreak +rr_cp_{max}\allowbreak +rp_{max}^2)$ and is comparable to the cost of reconstruction operators. Here the factors of $r_c\leq (N-r)$ can be further reduced if subleading singular values are further truncated. Notably, the DNA theory operates entirely in reduced-order space (both in modes and space), so the computational cost of predicting the entire risk curve does not scale with either the dimension $n$ or number $N$ of snapshots. By computing the ratio of leading-order terms, we get $T_{naive}/T_{DNA}= \order{nN/r}$. For the SST dataset this ratio is theoretically $\order{10^5}$, whereas our implementation in Python achieved a speedup of about $\order{10^3}$.

% \aak{check the uncertainty definition}
% Another related expression is the \emph{uncertainty heatmap}, or the predicted pixel-wise error bar of the reconstructed state extracted from the diagonal part of the covariance matrix:
% \begin{align}
%     \mathbf{\sigma}=&\sqrt{diag(\mathbf{K})}\label{eqn:heatmap}\\
%     \sigma_i=&\sqrt{K_{ii}}=\sqrt{\sum_{j}\mathbf{B}^2_{0,ij} + \sum_{j}\mathbf{B}^2_{1,ij} + \sum_{j}\mathbf{B}^2_{2,ij} + \sum_{j}\mathbf{B}^2_{3,ij}},
% \end{align}
% which is very similar to the expression in Ref.~\cite{klishin2023} which only included the sensor noise contribution $\mathbf{B}_3$.

\section{Case Study 1: Sparse Sensing}
\label{sec:sparsesensing}
\begin{figure*}
    \centering
    \includegraphics[width=.8\linewidth]{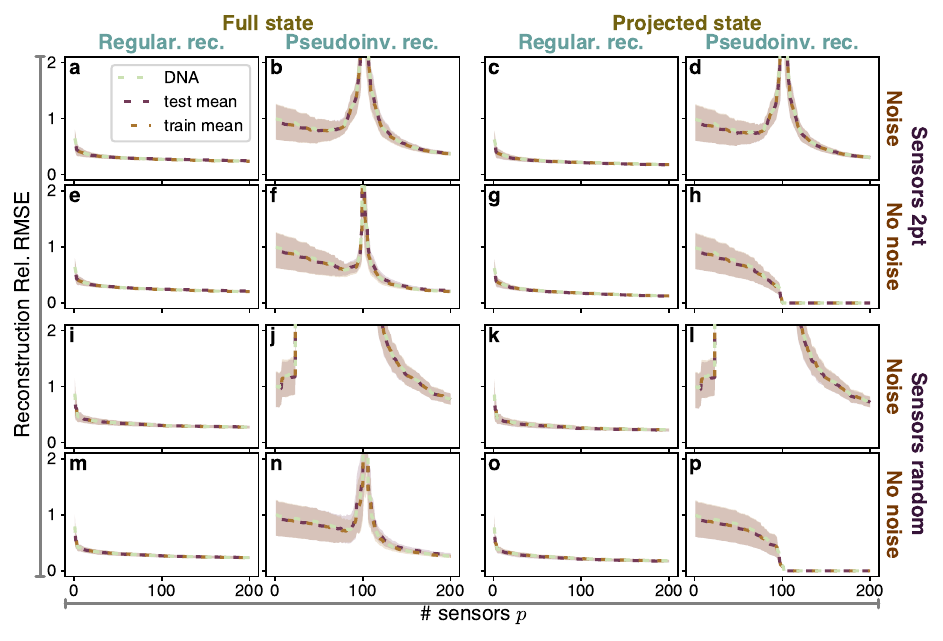}
    \caption{Comparison of benchmark reconstruction and the DNA theory prediction of risk curves of sparse sensing. Each design parameter of Fig.~\ref{fig:diagram} takes one of two values, resulting in $2^4=16$ combinations of all possible plots. The design choices are indicated on the labels of panel rows and columns. Each panel uses $r=100$ modes for reconstruction on the SST dataset and up to $p_{max}=200$ sensors.
    Dashed purple and brown curves and shaded areas show the empirical mean RMSE$\pm$one standard deviation, across test and train sets, respectively. Dashed green curve shows the analytic RMSE prediction with no free parameters.}
    \label{fig:factorial_analytic}
\end{figure*}

Following the theoretical derivation, we now turn to the first case study of sparse sensing on the SST dataset. 
We use the Eqn.~\ref{eqn:RMSE_DNA} to compute the analytical prediction for the same $2^4=16$ regimes tested in Fig.~\ref{fig:factorial}. We emphasize that due to the analytic averaging the Eqn.~\ref{eqn:RMSE_DNA} does not iterate over the individual states in the training and test sets, and thus the computation is dramatically faster, at about 3 seconds in a single-core process on a standard modern laptop as opposed to $\sim$50 minutes for the benchmark results in Fig.~\ref{fig:factorial} (speedup of a factor $10^3$).

The analytical curve is superimposed on top of benchmarking results in Fig.~\ref{fig:factorial_analytic}. Across all design parameters, the analytical curve precisely traces the empirical mean, reproducing not only the locations and magnitudes of the double descent spike, but also the minute localized jumps.

Having established the overall quantitative predictive power of the DNA theory, in the rest of this section we focus on how the individual terms $\mathbf{B}_{0,1,2,3}$ interact with the design factors to result in the leading mode reconstruction, the double descent instability peak and its connection to sensor placement, and the effect of regularization to suppress the double descent.

\subsection{Leading mode reconstruction}
We first consider the reconstruction of the leading modes, i.e. the $\mathbf{B}_1$ term in the covariance matrix \eqref{eqn:K}. In the simplest case of pseudoinverse reconstruction, $A=\boldsymbol{\Theta}^\dagger$, and the product $\boldsymbol{\Theta}^\dagger \boldsymbol{\Theta}$ is a projector onto the $Range(\boldsymbol{\Theta}^\top)$. When there are few sensors $p<r$, the matrix $\boldsymbol{\Theta}^\dagger \boldsymbol{\Theta}$ has exactly $p$ eigenvalues of 1 and $r-p$ eigenvalues of 0. When the number of sensors reaches $p\geq r$, if the sensing matrix $\boldsymbol{\Theta}$ remains full rank, the projector turns into identity and $\boldsymbol{\Theta}^\dagger \boldsymbol{\Theta}-\mathbb{I}_r=0$, and thus the $\mathbf{B}_1$ term ceases to contribute to the covariance matrix. If other terms do not contribute to the covariance either, the reconstruction becomes exact for any sensor selection (panels n,p in Fig.~\ref{fig:factorial_analytic}). This transition argument was also identified in GAD \cite{transtrum2025generalized} where the data insufficiency term turns to identical zero for more features than model parameters.

While the rank of the reconstruction increases linearly with the number of sensors as $\min(p,r)$, the error does not necessarily decrease linearly. While the leading $r$ modes carry the majority of signal variance, it is not uniform but is expressed by the singular values $\boldsymbol{\Sigma}'_r$. The projector $\boldsymbol{\Theta}^\dagger \boldsymbol{\Theta}$ does not project onto the first $p$ modes, but rather some linear subspace of the full $r$-dimensional space. The reconstructed state thus captures a progressively larger fraction of variance of the leading modes, but in a data-dependent way.

\subsection{Reconstruction instability}
\begin{figure*}
    \centering
    \includegraphics[width=.8\linewidth]{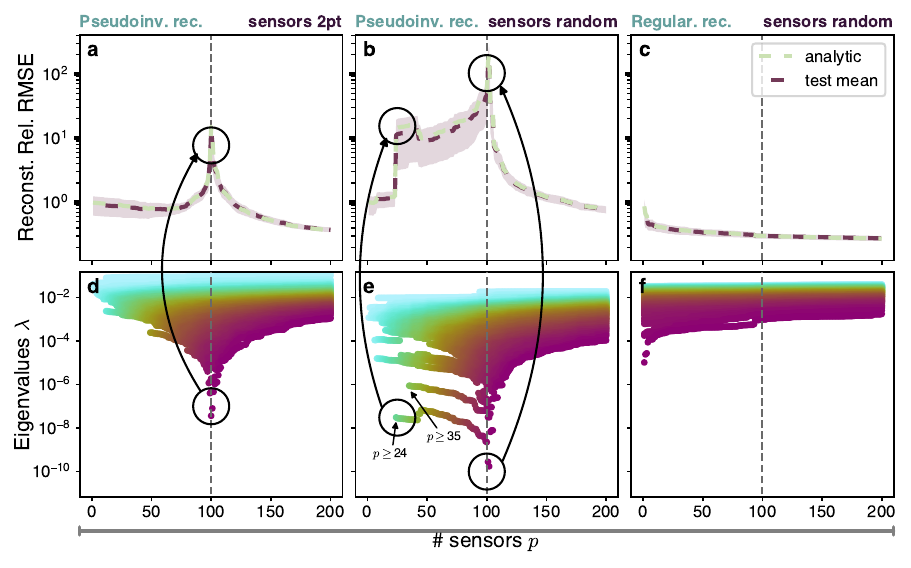}
    \caption{Reconstruction RMSE curves are tied to the spectrum of the reconstruction matrix. (a-c) RMSE curves for different regularization schemes and sensor selections with $r=100$ modes, both empirical and DNA prediction. (d-f) Spectrum of the reconstruction inversion matrix $\mathbf{M}$, with eigenvalues in ascending order by color. For the pseudoinverse reconstruction there are $\min(p,r)$ eigenvalues, for the regularized reconstruction there are always $r$ eigenvalues. Sharp spikes of RMSE curves precisely correspond to the emergence of low-lying reconstruction eigenvalues. For the 2-point method (d) the low-lying eigenvalues only appear at $p\approx r$, whereas for the random method (e) isolated low-lying eigenvalues appear earlier as marked.}
    \label{fig:spectral}
\end{figure*}
We now consider the last two terms of the covariance matrix $\mathbf{B}_2$ and $\mathbf{B}_3$ (Eqn.~\ref{eqn:B23}) in which the leading modes appear only once via the linear estimator $A$. These two terms are closely related if we define the following $p\times p$ matrices:
\begin{align}
    \mathbf{S}_{cont}=&\mathbf{C}\boldsymbol{\Psi}_c {\boldsymbol{\Sigma}'_c}^2 \boldsymbol{\Psi}_c^\top \mathbf{C}^\top \label{eqn:Scontamination}\\
    \mathbf{S}_{noise}=&\eta^2 \mathbb{I}_p \label{eqn:Snoise},
\end{align}
which quantify the covariance of two ``pathological signal'' inputs from the sensors into the reconstruction algorithm: $\mathbf{S}_{cont}$ quantifies the contamination of the signal by the subleading modes, whereas $\mathbf{S}_{noise}$ is the covariance of sensor measurement noise. In practical situations the magnitudes of these two matrices might be comparable, or one might be much larger than the other. However, so long as at least one of them is present, the spike in reconstruction error appears for number of sensors $p\approx r$ or even earlier Fig.~\ref{fig:factorial_analytic} panels j,l.

At the same time, the magnitude of the undesired inputs is typically much smaller than the magnitude of the leading modes, so they are somehow amplified. The only remaining component of the covariance term is the linear estimator matrix $\mathbf{A}$, the properties of which can be best understood through its singular value spectrum: so long as $\mathbf{A}$ has some large singular values, it would dramatically amplify the undesired input and lead to the spike. Consider a following more uniform definition of the reconstruction matrices:
\begin{align}
    \mathbf{A}_L=&\boldsymbol{\Theta}^\top \mathbf{M}_L^{-1}, \; p\leq r; \; &\mathbf{M}_L=\boldsymbol{\Theta}\boldsymbol{\Theta}^\top\\
     \mathbf{A}_L=& \mathbf{M}_L^{-1} \boldsymbol{\Theta}^\top,\;p> r;\; &\mathbf{M}_L=\boldsymbol{\Theta}^\top\boldsymbol{\Theta}\label{eqn:M_oversample}\\
     \mathbf{A}_R=&\mathbf{M}_R^{-1} \boldsymbol{\Theta}^\top,\;\text{any }p; \;&\mathbf{M}_R=\eta^2 {\boldsymbol{\Sigma}'_r}^{-2}+ \boldsymbol{\Theta}^\top\boldsymbol{\Theta},\label{eqn:MR}
\end{align}
from which it is clear that the matrix $\boldsymbol{\Theta}^\top$ appears identically in all expressions and cannot affect the spectrum.

We thus traced the spectrum of the reconstruction matrix $\mathbf{A}$ to the inversion of the matrix $\mathbf{M}$, the eigenvalues of which we directly contrast against the risk curve in Fig.~\ref{fig:spectral}. The eigenvalues span several orders of magnitude with the upper end of spectrum remaining nearly constant. Appearance of low-lying eigenvalues directly connects to the spikes in the risk curve: a small eigenvalue of $\mathbf{M}$ is inverted and contributes a large singular value to $\mathbf{A}$. 

\paragraph{Pseudoinverse reconstruction from optimized sensors}
In case of the 2-point sensor method (panels a,d), the lowest eigenvalue is around $10^{-7}$ and only appears at $p\approx r$, causing the relative RMSE to spike abruptly to around 10, i.e. the reconstruction error is 10 times larger than typical dataset variability. In the undersampled regime $p<r$, as the number of sensors increases, the number of eigenvalues of $\textbf{M}$ increases in sync, with the lowest eigenvalue gradually moving down. In the oversampled regime $p>r$, extra sensors contribute to the same number of eigenvalues $r$, and the lowest eigenvalue gradually increases.

\paragraph{Pseudoinverse reconstruction from random sensors}
In case of the random sensor placement (panels b,e), the first low-lying eigenvalue of about $10^{-8}$ appears abruptly at $p=24$, causing the relative RMSE to spike to 10. As the number of sensors grows further, a second distinct band of low-lying eigenvalues around $10^{-6}$ appears at $p=35$. At $p=r$, an even lower eigenvalue of $10^{-10}$ appears and the relative RMSE climbs to over 100, signifying a catastrophic reconstruction instability. In the oversampled regime $p>r$, the lowest eigenvalues rapidly increase, dropping the RMSE.

\paragraph{Regularized reconstruction from random sensors}
We now take the same random sensor set and apply the optimal regularized reconstruction (panels c,f). The lowest eigenvalue starts at around $10^{-5}$ but quickly increases as just a few sensors are added, reducing the relative RMSE under 0.5. Across the whole range of sensor numbers, low-lying eigenvalues never appear, resulting in a slow monotonic RMSE decrease. Since the inverted matrix $\mathbf{M}$ is regularized by the prior covariance, it is always well-conditioned for inversion.

In conclusion of this section, low-lying matrix $\mathbf{M}$ eigenvalues result in large singular values of the linear reconstruction matrix $\mathbf{A}$ and thus create the ``amplifier''. However, how are those eigenvalues connected to the individual sensors picked deliberately or randomly, and is the ``amplifier'' inevitable?

% Since the linear estimator $\mathbf{A}$ requires matrix inversion, we focus on the spectrum of the pre-inversion matrices and look for small eigenvalues:

% Let's first consider the pseudoinverse estimator $A_L=\boldsymbol{\Theta}^\dagger$ for the undersampled case $p<r$.

% also look at oversampled

% the estimator matrix contains an instability that would amplify the contamination or noise signal, so one of those is necessary

\subsection{Sensor placement}
\begin{figure*}
    \centering
    \includegraphics[width=.8\linewidth]{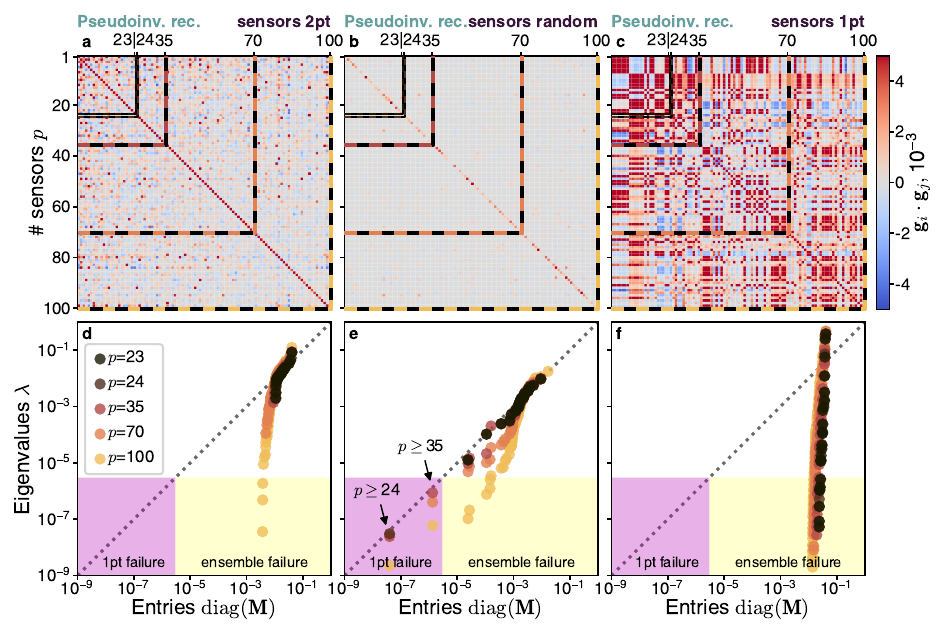}
    \caption{Sensor placement algorithms lead to either 1-point or ensemble failures of the sensor set. (a-c) Matrices $\mathbf{M}$ are formed by the leading principal submatrices of an ordered Gram matrix of sensing vectors; each entry is given by the dot product $\mathbf{g}_i\cdot \mathbf{g}_j$. Matrices follow the order of sensors placed by three different methods, and are limited by the dashed line border. The matrices for $p=23$ and $p=24$ are distinct, but the borders are right next to each other. (d-f) Ordered matrix eigenvalues plotted against the ordered diagonal entries. The 1-point failure is diagnosed by both eigenvalues and entries becoming low simultaneously, whereas the ensemble failure is diagnosed by only eigenvalues being low.}
    \label{fig:orthogonality}
\end{figure*}

Here we trace the spectral characterization of a set of sensors back to individual sensors (design factor 1). In the undersampled regime, the inversion matrix is $\mathbf{M}=\boldsymbol{\Theta\Theta}^\top$, yet $\boldsymbol{\Theta}=\mathbf{C}\boldsymbol{\Psi}_r$ is itself a row selection from the singular vector matrix. The singular vector matrix $\boldsymbol{\Psi}_r$ has orthogonal \emph{columns} per the POD construction (Eqn.~\ref{eqn:POD}), but $\mathbf{M}$ consists of dot products of \emph{rows}. Consider the singular vector matrix as a stacking of rows which we term \emph{sensing vectors} $\mathbf{g}_i\in \mathbb{R}^r$:
\begin{align}
    \boldsymbol{\Psi}_r=\mqty(-\mathbf{g}_1-\\-\mathbf{g}_2-\\ \vdots \\ -\mathbf{g}_n-).
\end{align}

%constructing the matrices and hypotheses
A matrix of \emph{all possible} dot products $\mathbf{g}_i \cdot \mathbf{g}_j$ would be the Gram matrix, yet we are interested in an \emph{ordered subset} of sensing vectors returned by a particular sensor placement algorithm. As the sensor budget increases, $\mathbf{M}:p\times p$ is formed by an ever-larger leading principal submatrix of the Gram matrix. As established in the previous section, reconstruction instability is directly tied to the emergence of small eigenvalues in the matrix $\textbf{M}$. We formulate two competing hypotheses for the emergence of small eigenvalues $\lambda$:
\begin{enumerate}
    \item \textbf{1-point failure hypothesis:} Small $\lambda$'s are driven by \emph{individual sensors}.
    \item \textbf{Ensemble failure hypothesis:} Small $\lambda$'s are driven by the \emph{collective sensor interactions}.
\end{enumerate}

%describe submatrices
In order to test the hypotheses, we plot the matrices $\textbf{M}$ in Fig.~\ref{fig:orthogonality}a-c following three sensor placement algorithms: the 2-point and random algorithms described above, and the 1-point algorithm of Ref.~\cite{klishin2023data} which selects for non-repeating locations of maximal signal variance but ignores sensor crosstalk (all illustrated in Fig.~\ref{fig:SST}b). The dashed lines in each panel delineate an ever-larger matrix $\mathbf{M}:p\times p$ with cuts at specific sensor counts $p$. The matrix for the 2-point method (panel a) is diagonally dominated but also shows a speckle of off-diagonal entries, both positive and negative. The matrix for the random method (panel b) shows a somewhat interrupted main diagonal and very infrequent off-diagonal entrees. The matrix for the 1-point method (panel c) shows rectangular regions of solid color, indicating a strong correlation between the sensors (both positive and negative). While these patterns are quite different, it is hard to intuitively relate them to the eigenvalue spectrum.

%naive approximation
Since the matrix $\mathbf{M}$ is formed by an outer product, it is symmetric, positive-definite and has all-positive eigenvalues. The diagonal part of the matrix $\mathbf{M}$ is also positive-definite. Since the first two matrices are diagonally-dominated, we propose the following deliberately na\"ive approximation of the spectrum:
\begin{align}
    \{\lambda(\mathbf{M})\}\approx \{ \diag(\mathbf{M})\},
    \label{eqn:spectrumdiag}
\end{align}
since for a diagonal matrix the eigenvalues are equal to its entries. If for a particular matrix this approximation happens to be good, then any emerging small eigenvalue is traced directly to an individual diagonal matrix entry and thus to an individual sensor, supporting the \textbf{1-point failure hypothesis}. On the contrary, if the approximation breaks down and small eigenvalues emerge \emph{without} small diagonal entries, then the small eigenvalues are driven by sensor interactions through the off-diagonal entries, supporting the \textbf{ensemble failure hypothesis}.

%test hypotheses
We test the hypothesis by plotting the two sides of Eqn.~\ref{eqn:spectrumdiag} against each other in Fig.~\ref{fig:orthogonality}d-f, where the \textbf{1-point failure} region is in the bottom-left corner and \textbf{ensemble failure} region is along the rest of the bottom edge. At each sensor budget, the number of eigenvalues equals to $p$. For the 2-point method (panel d), at small sensor budgets the eigenvalues and diagonal entries are both large and closely correlate, but as the budget gets larger, smaller eigenvalues emerge leading to an \textbf{ensemble failure} at $p=100$. For the random method (panel e), the eigenvalues and diagonal entries closely correlate for all but the largest sensor budgets. At the same time, the spectrum no longer forms a single continuous band: an isolated eigenvalue near $10^{-8}$ is absent at $p=23$ but appears at $p\geq 24$; another isolated eigenvalue appears at $p\geq 35$; finally, by $p=100$ the main band of eigenvalues also moves down. The random method thus exhibits an \textbf{early 1-point failure} as soon as a single bad sensor is placed, followed by a \textbf{collective failure}. For the 1-point method (panel f), all sensors are placed at locations of high signal variance, thus the diagonal entries are always large and eliminate the \textbf{1-point failure}; however, since the sensors are so correlated, the eigenvalue band spans the whole plotting range for any sensor budget $p$, and thus the \textbf{collective failure} also occurs immediately.

\paragraph{Inevitable orthogonality crisis}
Is the collective failure of sensors inevitable? Selecting a set of sensors is equivalent to picking out the sensing vectors $\mathbf{g}_i\in \mathbb{R}^r$. By the definition of the Euclidean space $\mathbb{R}^r$, there can be no more than $r$ linearly independent vectors. At the same time, the sensing vectors are not arbitrary but are selected from $n$ candidates. As each sensing vector typically has nonzero entries in all of the modes, it is impossible to select $r$ orthogonal vectors for any practical dataset. A ``smart'' sensor placement method (such at 2-point or QR) balances sensor variance and correlation and thus delays running out of orthogonal combinations to the last possible moment $p\approx r$; a ``dumb'' method (such as random) might stumble into low-variance individual sensing vectors or near-collinear sensing vectors early.

\paragraph{Oversampled regime}
While the orthogonality crisis is inevitable as $p$ approaches $r$ from below, the situation changes for $p>r$. Per Eqn.~\ref{eqn:M_oversample}, $p$ sensing vectors contribute to a fixed number $r$ of matrix eigenvalues. Each additional sensor adds a positive-definite contribution $\mathbf{g}_i \mathbf{g}_i^\top$ to the matrix $\mathbf{M}_L$, thus increasing the matrix values and driving up the whole eigenvalue spectrum \cite{transtrum2025generalized, bunch1978updating, wilkinson1965algebraic, thompson1976behavior, gantmacher2002oscillation}. Oversampling thus serves as an effective regularization of the reconstruction.

\paragraph{Sensor noise vs contamination}
The sensing vectors $\mathbf{g}_i$ express only the contribution of a particular location to the \emph{leading} $r$ singular vectors of the data, yet the locations also affect the \emph{subleading} spectral content as estimated in the contamination covariance Eqn.~\ref{eqn:Scontamination}.
%In empirical datasets, the same locations likely have high signal variance in leading and subleading modes. 
In case of no measurement noise, low contamination of certain sensor locations might not constitute a sufficiently large pathological signal to lead to a large error (e.g. Fig.~\ref{fig:factorial_analytic}n is smooth at $p\approx 24$), but adding a constant location-independent noise drastically increases the error (e.g. Fig.~\ref{fig:factorial_analytic}j has a jump at $p=24$).

% If contamination alone is present, certain low-contamination locations might not lead to a reconstruction instability, but adding noise 

% \aak{max p orthogonal vectors in Rp space}

% \aak{might get unlucky earlier, the smart method avoids this until it is not possible anymore}

% \aak{if the sensing vectors are nearly collinear, the reconstruction expects those two locations to carry the same signal in the first r modes. if the signal is not the same due to pathology, this blows up}

% \aak{pseudoinverse changes qualitatively for p>r, more sensors to contribute to same number of eigenvalues}

% \aak{also contamination and noise contribute differently
% contamination is location specific, explains panel n
% noise is location independent, hence leads to blow up}

%interpretation - what are you selecting for in different methods

%collective failure is inevitable - running out of orthogonality
%would not run out if regions were uncorrelated, but that signal would be incompressible

%noise and contamination are location specific

% how do we trace the emergence of low eigenvalues to individual sensors and is it inevitable?

% new figure on orthogonality

\subsection{Regularization}
\begin{figure}
    \centering
    \includegraphics[width=\linewidth]{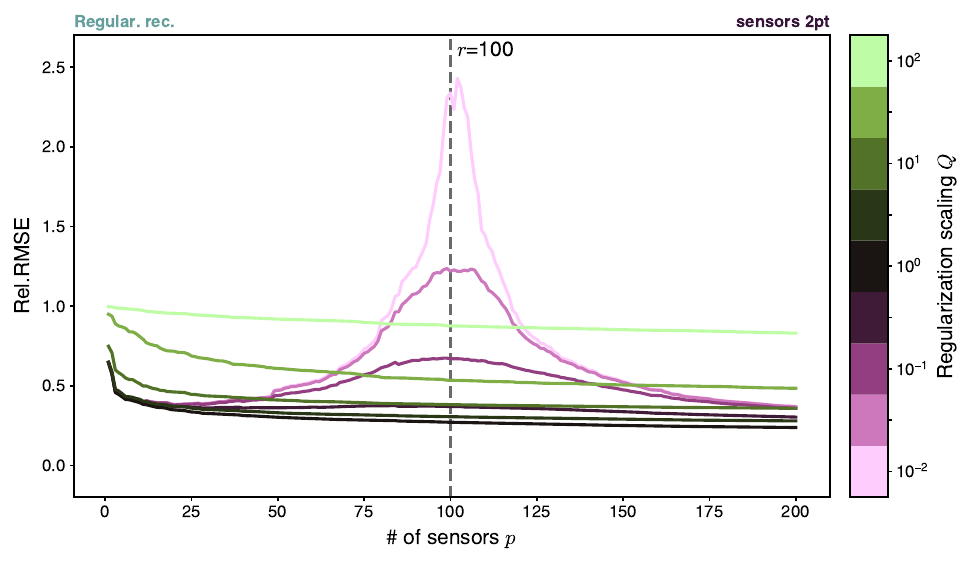}
    \caption{Effect of the regularization strength on the risk curves predicted by the DNA theory. Regularization scaling $Q$ is measured with respect to the optimal strength: purple curves reflect under-regularizing, green curves over-regularizing.}
    \label{fig:regularization}
\end{figure}

% \subsection{Why noiseless reconstruction can be exact}
% We first consider the conditions 

% \subsection{Why sensor placement matters}

% figure on collinear sensors

% \subsection{Why regularization helps}

% gets rid of low-lying eigenvalues

% \subsection{Why data truncation or projection helps}

% injected non-uniformly

% \subsection{Why noise hurts}

We now turn our attention to the effect of regularization. Qualitatively, regularization appears as adding a positive diagonal contribution to the matrix $M_R$ in Eqn.~\ref{eqn:MR}. Since in the previous section we directly traced the reconstruction instability to the appearance of low-lying eigenvalues in the $M$ matrix, addition of of a positive diagonal contribution raises the eigenvalues, evident in the spectrum in Fig.~\ref{fig:spectral}f. The regularization matrix $\eta^2 {\boldsymbol{\Sigma}'_r}^{-2}$ is directly computed from the singular values of the training data and is thus not isotropic: the first few modes are regularized much less than the later ones. At the same time, the expected noise magnitude $\eta^2$ also scales the expression. If the noise is large, the regularization becomes stronger, and thus the reconstruction effectively ``falls back'' onto the prior; if the noise is small and sensors are precise, the regularization gets weaker and sensor readings are more important for reconstruction.

Since the ratio of expected noise and data variance seems to affect the optimal regularization strength, we ask a more general question: how does the risk curve shape change with regularization strength? Consider the following generalized reconstruction matrix:
\begin{align}
    \mathbf{A}_Q=(Q \eta^2 {\boldsymbol{\Sigma}'_r}^{-2}+\boldsymbol{\Theta}^\top \boldsymbol{\Theta})^{-1} \boldsymbol{\Theta}^\top,
\end{align}
where $Q$ is an auxiliary scaling factor. For $Q=1$, $\mathbf{A}_Q=\mathbf{A}_R$ and the ``optimal'' linear estimator is reproduced, but in general both higher and lower values can be considered. We present a sweep over the regularization scaling in Fig.~\ref{fig:regularization} where all other reconstruction parameters are kept fixed. The $Q=1$ curve (black) serves as an envelope for all other curves as it is the lowest. Under-regularizing $Q<1$ (purple) results in a progressively steeper peak at $p\approx r$ as the reconstruction matrix approaches the pseudoinverse $\mathbf{A}_L$. Over-regularizing $Q>1$ results in monotonic curves that gradually get higher. Getting the regularization strength of the correct magnitude is thus important for reconstruction, but the exact value is less crucial.

% strength of regularization is ratio of expected noise and prior
% what if expected noise mismatches the actual noise?

% also does not take the contamination into account

% \aak{noise/perturbation injected uniformly, amplified in resonance}

% \aak{plot fpr over/under regularization}

% \section{Reconstruction design}
% \subsection{Optimal estimators}
% depends on the exact statistical model assumed for the signal
% flat prior, only noise, noise and contamination
% wrong choices/unmodeled noise lead to attempts to overfit

% \subsection{Optimal number of sensors}

\section{Case study 2: Empirical interpolation for PDEs}
\label{sec:DEIM}

The second case study is the Discrete Empirical Interpolation Method (DEIM) \cite{chaturantabut2009discrete, drmac2016qdeim}, a data-driven method of constructing ROMs for PDEs. Similarly to sensor placement, DEIM relies on a tailored basis to reconstruct high-dimensional vectors from sparse measurements. However, while for sensor placement computing the reconstruction is the end of story, for DEIM the reconstruction is inserted into a time stepping algorithm to compute approximate numerical time-dependent solutions.

In this section, we start with briefly reviewing the canonical DEIM construction and disambiguating several method parameters that are commonly considered identical. We then introduce our case study, the breather solutions in the Nonlinear Schr\"odinger Equation (NSE), and the required modifications of the sparse sensing method for the complex-valued quantities. We show how the DNA theory applies to risk curves of static empirical interpolation (instantaneous state) and trace its effects on the spatiotemporal numerical solutions. Our construction of DEIM and the specific case study follows Chapters 12-13 of Ref.~\cite{brunton2022databook}.

\subsection{DEIM method}

Consider a prototypical nonlinear PDE:
\begin{align}
    \partial_t\mathbf{u} = \mathbf{Lu}+\mathbf{N}(\mathbf{u}),
    \label{eqn:PDE}
\end{align}
where $\mathbf{u}$ is a spatiotemporal field, $\mathbf{L}$ is a \emph{linear} operator (commonly containing spatial derivatives), and $\mathbf{N}(\cdot)$ is a \emph{nonlinear} operator (commonly acting locally). If the field is discretized onto a grid or mesh of size $n$, $\mathbf{u}\in \mathbb{R}^n$ or $\mathbf{u}\in \mathbb{C}^n$ (depending on the PDE considered). The linear operator $\mathbf{L}$ is then an $n\times n$ matrix, and the nonlinear operator $\mathbf{N}(\cdot)$ needs to be evaluated on $n$ inputs. The goal of DEIM is to use the prior knowledge of high-fidelity solutions to construct a ROM. Since a DEIM approximation needs to be evaluated in a time integration loop, the goal of its construction is to avoid any $\order{n}$ operations performed \emph{repeatedly} (but they can be precomputed once before time integration).

Let $\mathbf{X}:n\times N$ denote the training data obtained by full-order integration (stacking the $\mathbf{u}$ vectors) and $\mathbf{X}^{NL}=\mathbf{N}(\mathbf{X}):n\times N$ denote the nonlinear term evaluated on each training snapshot. Model reduction in DEIM has the same two flavors of mode-sparsity and space-sparsity: the system state $\mathbf{u}$ and the nonlinear term $\mathbf{N}(\mathbf{u})$ are approximated with two separate finite numbers of modes, and the evaluation of the local nonlinear term is confined to a few locations. While in conventional DEIM the three sparsity parameters are often taken to be the same, here we separate them explicitly:
\begin{itemize}
    \item $r$ is the number of modes used to approximate the system state $\mathbf{u}$;
    \item $q$ is the number of modes used to approximate the nonlinear term $\mathbf{N}(\mathbf{u})$;
    \item $p$ is the number of sensors (interpolation points) used for the nonlinear term.
\end{itemize}

The problem of reconstructing $q$ modes of the nonlinear term from $p$ localized measurement reduces to sparse sensing covered in Case 1, but the need to perform time integration of $r$ modes enriches the problem.

\subsubsection{DEIM modes and reconstruction}
The tailored modes for the state itself and the nonlinear reconstruction are found by two separate PODs:
\begin{align}
    \mathbf{X}=&\mathbf{U}\boldsymbol{\Sigma}\mathbf{V}^\top\approx \mathbf{U}_r \boldsymbol{\Sigma}_r \mathbf{V}_r^\top \\
    \mathbf{X}^{NL}=&\boldsymbol{\Xi}\boldsymbol{\Sigma}^{NL}\mathbf{V}^{NL,\top} \approx \boldsymbol{\Xi}_q\boldsymbol{\Sigma}_q^{NL}\mathbf{V}_q^{NL,\top},
\end{align}
where the subscripts $r,q$ denote truncation to the specified number of modes. Since sparse sensing is only applied to the nonlinear term, we only need to normalize the singular values ${\boldsymbol{\Sigma}'_q}^{NL},{\boldsymbol{\Sigma}'_c}^{NL}$ (leading and subleading modes, respectively).

The system state is approximated with $\mathbf{u}\approx \mathbf{U}_r\mathbf{a}_r$, where $\mathbf{a}_r$ are time-dependent mode coefficients. With this low-rank Galerkin approximation, the whole PDE of Eqn.~\ref{eqn:PDE} is reduced to a dynamical system of dimension $r\ll n$. In order to construct the ROM equations, we need to approximate the nonlinear term with sparse measurements. For a selection matrix $\mathbf{C}$ the nonlinear term \emph{at sensed locations} is evaluated by $\mathbf{N}(\mathbf{C}\mathbf{U}_r \mathbf{a}_r)$. From these sparse ``measurements'' of the nonlinearity, we estimate the whole field of the nonlinear terms with sparse sensing methods.

We construct the ROM equation by: (i) projecting the full Eqn.~\ref{eqn:PDE} onto $\mathbf{U}_r$ modes and (ii) replacing the full-field nonlinear term with a sparse sensing version. The resulting dynamical equation is as follows:
\begin{align}
    \partial_t \mathbf{a}_r = \underline{\mathbf{U}_r^\top \mathbf{L} \mathbf{U}_r} \mathbf{a}_r+\underline{\mathbf{U}_r^\top \boldsymbol{\Xi}_q \mathbf{A}}\mathbf{N}(\underline{\mathbf{C} \mathbf{U}_r}\mathbf{a}_r), \label{eqn:DEIM}
\end{align}
where the underlined expressions are matrix products that involve $\order{n}$ operations but can be \emph{precomputed before time-stepping}. In the first term, $\mathbf{U}_r^\top \mathbf{L} \mathbf{U}_r$ accounts for the effects of the linear operator (such as spatial derivatives) on the field modes. In the second term, $\mathbf{U}_r^\top \boldsymbol{\Xi}_q \mathbf{A}$ computes a reconstruction of nonlinear mode coefficients from sparse measurements with one of the previously introduced estimators $\mathbf{A}$ (pseudoinverse or regularized), then reconstructs full-resolution nonlinear term and projects it onto the state modes. Inside the bracket, $\mathbf{C} \mathbf{U}_r$ approximates the state values at the sensing locations. Note that due to a finite number of modes $r$, the approximated state is slightly different from the true state, and thus leads to an error in the sparsely sensed nonlinear term; this serves as an effective measurement ``noise'', the covariance of which is hard to characterize in practice.

% sigma scale and sigma scale NL
% contamination

% general pde - linear and nonlinear operators

% nonlinear is local, so reducing the evaluations of that is the goal

% place where we generalize is twofold - separate r, q, p and complex

% adaptive contamination and effective noise

\subsubsection{Complex-valued corrections}
\begin{figure*}[t!]
    \centering
    \includegraphics[width=.8\linewidth]{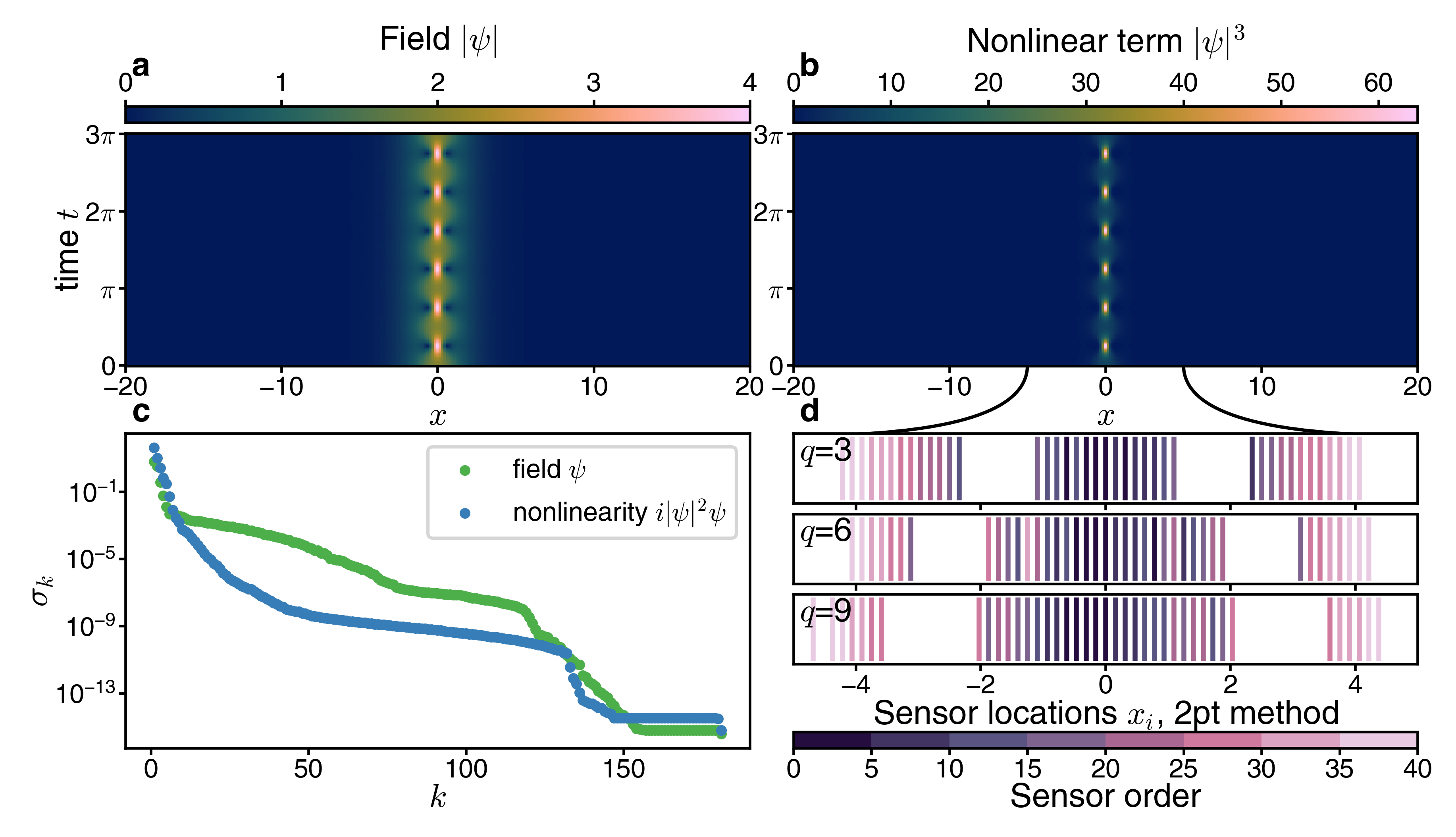}
    \caption{Training dataset and sensor placement for the Nonlinear Schr\"odinger Equation case study. (a) Spatiotemporal pattern of the absolute value of the complex field $\abs{\psi}$ with $\pi/2$ periodicity in time. (b) Spatiotemporal pattern of the absolute values of only the nonlinear term $\abs{\psi}^3$. (c) Singular value decay for both datasets. (d) Sensor placement at different nonlinear mode numbers $q$, color coded by placement order.}
    \label{fig:DEIM_data}
\end{figure*}
Additional care in DEIM construction needs to be taken to account for complex-valued fields used in some PDEs. If the state $\mathbf{u}$ is complex valued, so are usually the state and nonlinearity data matrices $\mathbf{X},\mathbf{X}^{NL}$. POD can still be straightforwardly computed for a complex-valued matrix. The resulting singular \emph{values} are real, as the complex phase can be moved to the singular \emph{vectors}.

If the true state $\mathbf{u}$ and some approximation or estimation $\hat{\mathbf{u}}$ are available, the difference between them can still be quantified by relative RMSE defined in Eqn.~\ref{eqn:RMSE}, where the vector norm operation maps complex vectors to real non-negative norm values. The reference scale of the data matrices $\sigma_{scale},\sigma_{scale}^{NL}$ can be estimated as the (real) standard deviation of the corresponding matrices. Similarly, the DNA prediction of RMSE is computed following Eqn.~\ref{eqn:RMSE_DNA} using the absolute values of squares of the $\mathbf{B}$ matrix elements.

The pseudoinverse reconstruction matrix $\mathbf{A}_L= \boldsymbol{\Theta}^\dagger$ is still well-defined for complex $\boldsymbol{\Theta}$. The regularized reconstruction is updated by replacing the matrix transpose with the conjugate transpose $\boldsymbol{\Theta}^\top\to \boldsymbol{\Theta}^*$ to yield the reconstruction matrix:
\begin{align}
    \mathbf{A}_R=(\eta^2_{reg}{\mathbf{\Sigma}'_q}^{-2}+ \boldsymbol{\Theta}^*\boldsymbol{\Theta})^{-1} \boldsymbol{\Theta}^*,
\end{align}
where the inverted matrix in the bracket is Hermitian by construction and thus has all real eigenvalues.

The regularization parameter $\eta_{reg}$ needs to be selected in anticipation of the expected mismatch between the sampled and true values of the nonlinear term. Given that there is no true measurement noise in a simulation, the main contribution would come from the contamination by residual modes. We thus assign the regularization parameter for a $q$-mode approximation as follows:
\begin{align}
    \eta^2_{reg}=\frac{1}{nN}\sum\limits_{i=q+1}^{max}\sigma_i^2,
    \label{eqn:etareg}
\end{align}
where the sum is taken over all subleading nonzero singular values.

% complex generalization for RMSE and optimal reconstruction
% eqn:RMSE still works, norm is understood to be complex (square and absolute value)

\subsection{Nonlinear Schr\"odinger equation}
\begin{figure*}
    \centering
    \includegraphics[width=.8\linewidth]{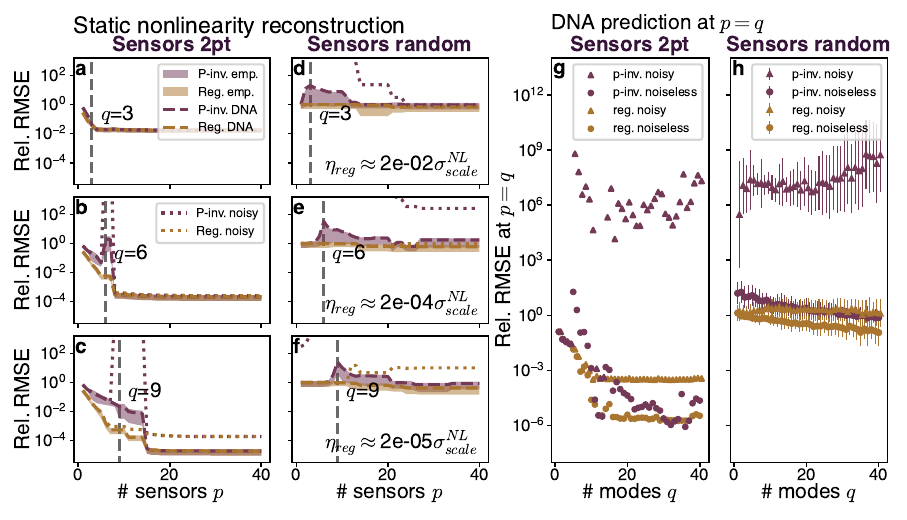}
    \caption{Statistics of static reconstruction of the nonlinear term. (a-f) Risk curves for pseudoinverse (``p-inv.'') and regularized (``reg.'') reconstructions across the empirical snapshots (shaded region of 1 std) and DNA prediction (dashed lines). Dotted lines show the DNA risk curve under addition of observation noise of magnitude $10^{-3}$. Rows correspond to different mode numbers $q$, columns to 2pt and random sensor placement. Regularization strength $\eta_{reg}$ is adaptive for each $q$ level as shown in (d-f). (g-h) DNA prediction for the worst-case risk scenario $p=q$ across sensor placement (columns), pseudouinverse and regularized reconstruction (purple and orange), and noiseless and noisy measurements (circles and triangles, all noisy measurement markers are shifter to the right by $\Delta q=0.5$ for visual clarity). For random sensors (h) the error bar shows the mean and standard deviation log-averaged over 50 random sensor set replicas.}
    \label{fig:DEIM_static}
\end{figure*}

In order to illustrate the range of DEIM behaviors, we consider a specific example of the Nonlinear Schr\"odinger Equation (NSE):
\begin{align}
    \partial_t \psi=\frac{i}{2}\partial_{xx}\psi+i\psi\abs{\psi}^2,
\end{align}
where $\psi(x,t)$ is a complex-valued wavefunction defined on dimensionless 1D space $x$ and time $t$. The first term expresses the linear part of dynamics, the second term the nonlinear part. We specifically consider the initial condition $\psi(x,0)=2/\cosh{x}$, which is known to result in a breather solution, a nonlinear spatiotemporal oscillation with a time period of $\pi/2$.

We use a numerical solution of the NSE to generate training data for DEIM. We define the spatial domain $x\in[-20,20]$ discretized into $n=256$ equally spaced points. The equation is solved in spatial Fourier domain:
\begin{align}
    \partial_t \tilde{\psi}=-i\frac{1}{2}k^2 \tilde{\psi} + i \mathcal{F}(\psi \abs{\psi}^2),
\end{align}
where $\tilde{\psi}(k,t)=\mathcal{F}(\psi(x,t))$ is the frequency parameterized wavefunction,  $\mathcal{F}$ and $\mathcal{F}^{-1}$ are forward and inverse Fourier transforms, and $k$ are the Fourier wavenumbers. Wavenumber multiplication replaces spatial differentiation in Fourier space. The resulting equation is treated as a system of 256 coupled ODEs for each spatial Fourier mode and integrated numerically with the 4th order Runge-Kutta method on the time domain $t\in [0,3\pi]$ (six oscillation periods) with 180 time steps ($\Delta t\approx 0.052$). It is crucial to collect data for an integer number of oscillation periods so that all phases of the oscillation are equally represented and the first few modes follow the spatial symmetries and contain most of data variance. The resulting time series is inverse Fourier transformed, and the snapshots are stacked into the training data matrix $\mathbf{X}$ (Fig.~\ref{fig:DEIM_data}a). The nonlinear term data matrix is obtained by computing $i\mathbf{X}\abs{\mathbf{X}}^2$ elementwise (Fig.~\ref{fig:DEIM_data}b).

The data matrices clearly demonstrate the periodic spatiotemporal oscillations confined mostly to the small region near $x=0$. The wide spatial domain is important to resolve high spatial frequencies with the universal Fourier basis and achieve a numerically stable solution, even though the majority of space has virtually no dynamics. Both the field itself and the nonlinear term demonstrate a sharp decay of singular values (Fig.~\ref{fig:DEIM_data}c), motivating constructing a ROM in a tailored basis. We use the same 2-point sensor placement algorithm to place up to 40 sensors. Since the algorithm is sensitive to the number of modes $q$, we show representative sensor placements color-coded by order in Fig.~\ref{fig:DEIM_data}d, where all sensors are confined to the central region $x\in[-5,5]$ and only differ slightly in order depending on $q$. In contrast, the random sensor placement will mostly place sensors outside of this central region in locations of very low signal.

% breathers

% benchmark solution via fourier

% RK4 integration with 180 time steps

\subsection{Reconstructing the nonlinear term}
\begin{figure*}[ht]
    \centering
    \includegraphics[width=.8\linewidth]{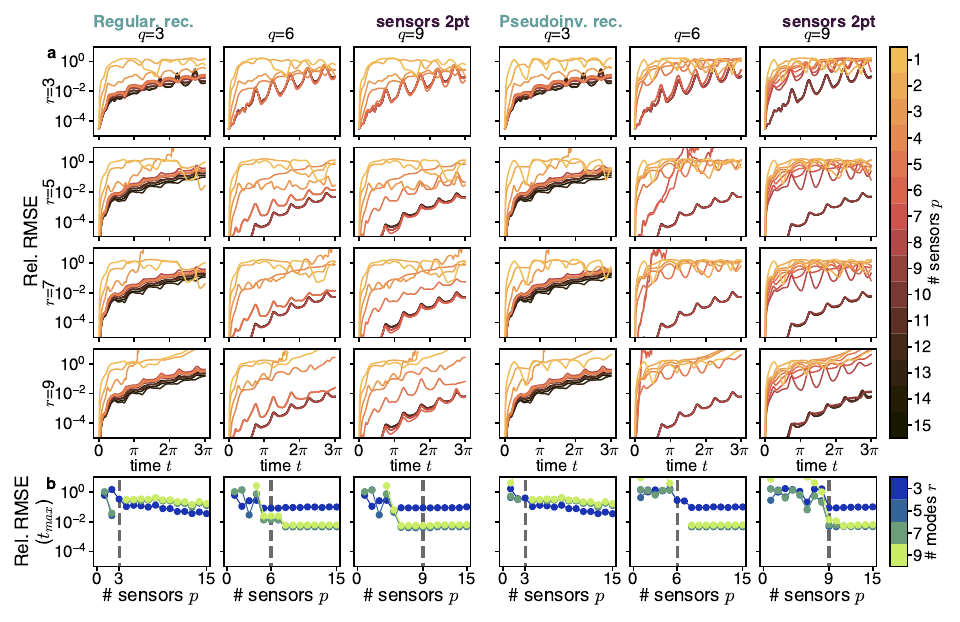}
    \caption{Empirical error curves of time-integrated DEIM. (a) Time-dependent error curves across different numbers of $r$ modes (rows), $q$ modes (columns), sensor numbers (curve colors), and reconstruction methods (left and right halves). Wavy curve patterns correspond to the ground truth time period of $\pi/2$. Time integration is set to terminate at $\norm{\mathbf{a}_r}>10^3$ to avoid numerical divergence. (b) Empirical error at the final time point $t_{max}=3\pi$ across sensor numbers and $r$ modes (colors). $q$ modes and reconstruction methods same as (a).}
    \label{fig:DEIM_trajectories}
\end{figure*}

We now proceed with constructing DEIM models at different levels of approximation. We start with analyzing a static reconstruction of the nonlinear term from sparse observations. Specifically, for $p$ local observations of the nonlinear term and $q$ reconstruction modes, what is the level of error as measured by relative RMSE? Reconstruction error here \emph{does not depend} on the number of modes $r$ as there is no time integration. In addition, since the ground truth data is periodic (up to the integration error of the full-order model), we do not separate the data into train and test sets as they would be identical. We consider both the noiseless case (the nonlinear term is observed exactly but sparsely, both empirically and in the DNA theory) and the case of uncorrelated noise of relative magnitude $\eta_{noise}/\sigma_{scale}^{NL}=10^{-3}$ to emulate the time integration error (only in the DNA theory).

RMSE patterns for the nonlinear term of NSE broadly resemble those from Case study 1 (Fig.~\ref{fig:DEIM_static}). For 2-point sensors (a-c), the regularized reconstruction produces a risk curve that monotonically decreases until a $q$-dependent saturation level. The pseudoinverse reconstruction shows a slightly different risk curve: at $p=q=3$ it matches the regularized curve; at $p=q=6$ a peak appears since the inversion matrix is large enough to have low eigenvalues but the contamination magnitude $\eta_{reg}\approx 2\cdot 10^{-4}\sigma_{scale}^{NL}$ is still appreciable; at $q=9$ the whole curve separates from the regularized one and decays much slower until $p\approx 15$. For random sensors (d-f) the risk curve is qualitatively different: the regularized curve remains nearly flat at Rel.RMSE$\approx 10^0$ and the pseudoinverse reconstruction is systematically worse with a pronounced peak at $p=q$. Across all of those cases, the DNA theory closely follows the empirical curve. The addition of noise (dotted curves) for the 2-point sensor case merely emphasizes the $p=q$ peak; for the random case the pseudoinverse curve increases by orders of magnitude and mostly moves above the axes range. The regularized curve depends on the adaptive regularization strength $\eta_{reg}$: for $q=3$ the regularization is stronger than noise level and thus the noisy reconstruction curve matches the noiseless one; for $q=6,9$ the regularization is weaker than noise and thus at $p>q$ the noisy curve lies above the noiseless one by an order of magnitude or more.

The shape of the risk curves suggests that the worst case scenario is choosing $p=q$ (peak of the risk curve), so we explore the height of that peak further with the DNA theory (Fig.~\ref{fig:DEIM_static}g-h). For 2-point sensors, all reconstruction regimes overlap for $q\leq 4$ but show a drastic difference at $q>4$. The regularized reconstruction errors decay nearly monotonically, reaching different asymptotic levels for the noiseless and noisy cases. The pseudoinverse reconstruction shows a different pattern: the noiseless error briefly spikes to $10^1$ at $q=5$ and then follows an irregular decay, while the noisy error jumps up by orders of magnitude to around $10^6$ and stays there.
%For random sensors, we average the DNA predictions over sensor realizations, with regularized reconstruction error remaining nearly flat at $10^0$, noiseless pseudoinverse error at a similar level, and noise pseudoinverse error around $10^6$.
%The graphs of error in the noisy case should only be interpreted in a qualitative sense, since the ``noise'' originates in time integration from the $r$-mode approximation of the state, passed through the nonlinear operator $\mathbf{N}(\cdot)$, and evaluated only at specific sensor locations. However, even with those complicating factors, 
Fig.~\ref{fig:DEIM_static}h directly proves that random sensors have no chance of generating accurate DEIM predictions: an approximation of a differential equation with a relative error of the right hand side consistently higher than $10^0$ cannot possibly result in a correct trajectory. In the following section on time integration results, we restrict consideration to \emph{only 2-point sensors} (though other placement methods such as QR pivoting \cite{drmac2016qdeim, manohar2018data} can also select much-better-than-random sensor sets).

% first consider the nonlinear term, so ignore r
% several q values and continuous p. empirical and risk curves, no division of test and training as they are identical, prediction works
% at q=6 see a peak - there is pathological signal and amplification

% adding fake noise - strong double descent peak at p=q

% now want to know more broadly how bad RMSE can get at p=q
% for good sensors there is a steady decay to the level of 10-5, but unregularized noise is catastrophic. 10-3 noise is a lot given the contamination scale
% for random sensors average is at 10-0 so the sensing is essentially useless

\subsection{Time integration}

We proceed with analyzing the error of time-integrated DEIM (Eqn.~\ref{eqn:DEIM}) using 2-point sensors. The broad question is whether approximating the differential equation with small $r,q,p$ leads to a good approximation of the trajectory. The initial condition $\psi(x,0)$ is projected onto the $r$ modes used for each set of parameters. We again use the 4th order Runge-Kutta method with 180 time steps over $t\in[0,3\pi]$ so that the original time series and the DEIM approximation have the same domain and are directly comparable. The time dynamics are reduced to an $r$-dimensional dynamical system, the qualitative properties of which (stability, oscillations, boundedness) are hard to quantify \emph{a priori}. Empirically, the error grows over time with some fluctuations owing to the periodic nature of the ground truth dynamics (Fig.~\ref{fig:DEIM_trajectories}a). The case of $r=q=p=3$ from Ref.~\cite{brunton2022databook} is contained in the top-left panel as a special case.

The general observation across the panels is that as the number of sensors $p$ increases, the error curve approaches an asymptotic form for given $r,q$. The asymptotic error drops almost an order of magnitude from $r=3$ to $r=5$ but remains consistent afterwards, suggesting that a sufficient part of dynamics has been resolved by $r\geq 5$ modes. The number of sensors required to reach the asymptote can be gauged by the color of the curve (lowest $p$ value), with distinctly fewer sensors required with regularized reconstruction. Another distinct feature is runaway dynamics with RMSE diverging off beyond the plotting range, only observed for many modes $r>3$ and close to the reconstruction instability point $p\approx q$. The runaway behavior occurs much earlier in time for the pseudoinverse reconstruction at $q=6$.

In order to summarize the error patterns and identify the double descent signature, Fig.~\ref{fig:DEIM_trajectories}b plots only the final value of the error at $t_{max}=3\pi$ across the sensor number. For $q=3,6$ one or several data points are missing at or near $p=q$, owing to the divergence of numerical integration. At $q=3$ the error curve shows a slow unsteady decay, whereas for $q=6,9$ there are distinct plateaus at a high number of sensors, with the regularized reconstruction arriving at the plateau with smaller $p$.

% completely exclude ranom sensors, 2pt only

% now reconstruction - time series of error
% p=r=q=3 case is in top-left panel

% with sufficiently many sensors can get to fairly low error - but what is the asymptote and how fast to reach it? graphically, what is the colro of the asymptote = lowest p for which it is reached. definitely lighter for regularized

% bigger r allows for a more complex dynamical system such as runaway (not seen for r=3). maybe it is possible to construct ROM equations that only support a limit cycle

% overall trend - b left has undefined behaviors at p=q
% regularization makes the curve more monotonic and reach the asymptote better

% \aak{r modes for state, q modes for nonlinearity, p sensors}

% \aak{using p=r is the worst possible choice}

\section{Discussion}
\label{sec:disc}
\begin{figure}
    \centering
    \includegraphics[width=\linewidth]{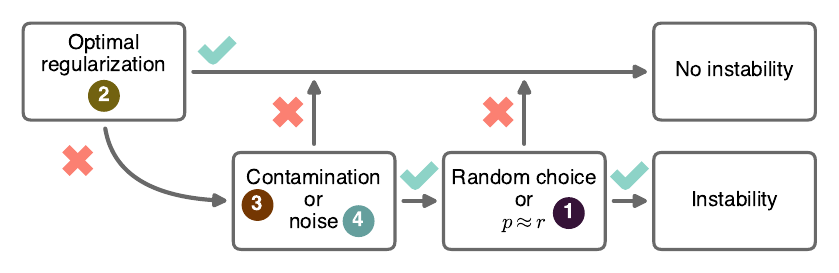}
    \caption{Flowchart of sufficient reasons for reconstruction instability to emerge out of design choices. When available, an optimal regularization scheme (2) entirely suppresses the instability. Otherwise, presence of high-mode contamination (3) or measurement noise (4) creates a pathological signal fed into the ``amplifier''. If the sensor choice is either random or has run out of viable orthogonal options (1), the ``amplifier'' acts on the pathological signal and creates an instability.}
    \label{fig:flowchart}
\end{figure}

% \aak{fill this in}
% \aak{cite exposure theory, mention again the mode-sparsity and space-sparsity}

\subsection{Origins of double descent}
In this paper we investigated the origins of the double descent phenomenon, or an abrupt spike in reconstruction error, in context of static and dynamic ROMs. In this context, the reconstruction instability directly results from the inherent tension between mode-sparsity and space-sparsity, as opposed to the training set size in studies of double descent in supervised learning \cite{belkin2019reconciling, nakkiran2021deep, rocks2022memorizing}.

Quantitatively, we proposed the Data-Noise Averaging theory to create a computationally cheap yet accurate approximation of the reconstruction error covariance matrix for \emph{any linear reconstruction setup and any sensor set}. The DNA theory approach of preserving system-specific quantities (singular vectors and values) while averaging out realization-specific ones (test states and noise) is similar to the exposure theory for learning complex network structure (preserved) with random walks (averaged-out) \cite{klishin2022exposure, klishin2022learning}. The key DNA Eqn.~\ref{eqn:K} shows that the covariance error includes four terms of which only the noise-driven one was accounted for in our prior work \cite{klishin2023data}; these terms have direct parallels to the GAD theory \cite{transtrum2025generalized} yet are used here as exact quantities rather than merely upper bounds in GAD. As a result, the DNA theory accurately predicts the risk curves across a variety of empirically-tested sparse sensing scenarios throughout the paper and can be used to test alternative, counterfactual ROM setups.

% The resulting risk curves decouple the understanding and prediction of reconstruction error from design and optimization of a sensing setup. The key Eqn.~\ref{eqn:K} shows that the covariance error includes four terms, of which only one was accounted for in our prior work \cite{klishin2023}. \emph{Quantitatively}, the covariance expression is used to predict the risk curves across a variety of sparse sensing scenarios throughout the paper.

Beyond the numerical values, \emph{qualitatively} the covariance expression allows answering the questions of what causes double descent and how to prevent or at least mitigate it. We summarize the answer to these questions in the flowchart of Fig.~\ref{fig:flowchart}. Optimal regularization of the reconstruction, if available, guides the reconstructed state to lie within a prior learned from the data, and thus entirely suppresses the reconstruction instability. If the regularization is unavailable, the reconstruction can be thought of as a ``pathological signal'' entering an ``amplifier'': the instability arises if both are present. The pathological signal is anything not contained in the leading $r$ modes, including any combination of measurement noise and contamination from subleading modes. The amplifier originates from low-lying eigenvalues in unregularized matrix inversion which originate either from unfortunate random sensor choice or from the orthogonality crisis in sensing vectors in $\mathbb{R}^r$.

% linear reconstruction can be easily updated for any desired sensor set

% flowchart summary

% which specific sensors to blame. analytic theory makes exploration dramatically faster, e.g. rescale regularization or try other sensor placements

\subsection{Reconstruction design}
Sparse sensing, like many other linear reduced-order models, has been based on the POD of the dataset (Eqn.~\ref{eqn:POD}). POD decomposes the data into \emph{three} matrices and previous sparse sensing approaches only used the \emph{first} matrix of singular vectors to construct a data-driven basis \cite{chaturantabut2009discrete, drmac2016qdeim, manohar2018data}. The regularized reconstruction introduced independently in Refs.~\cite{klishin2023data, Kakasenko2026BridgingTG} also uses the \emph{second} matrix of singular values to parametrize the Bayesian prior for a regularized reconstruction. The more knowledge is available about the reconstruction setup, the more it can be exploited to control the error.

The strength of regularization is given by the \emph{ratio} of noise and prior covariances: if sensors are precise, their readings are given more weight; if sensors are noisy, reconstruction falls back onto the prior. The reconstruction $\mathbf{A}_R$ used in this paper assumes that \emph{noise dominates over the contamination}, as would be common in experimental measurements. On the contrary, in a high-fidelity numerical simulation the value of a given pixel can be read off exactly, so contamination would dominate and necessitate a different estimation of regularization strength akin to our Eqn.~\ref{eqn:etareg}. At the same time, in a low-fidelity numerical simulation, as seen in our DEIM model, errors compound throughout the time integration and lead to an effective ``noise'' in evaluating the right hand side of the equation.

The regularized reconstruction is in general desirable and sufficient to avoid instability, but it is not always available. Non-POD methods might provide a good reconstruction basis but no expectation of variance in each mode and thus no way to construct a prior. The pseudoinverse reconstruction $\mathbf{A}_L$ remains available in that case, but as we argue in this paper it suffers from a generic instability at $p\approx r$ for good placement methods or even earlier for bad ones. In other words, the $p=r$ convention motivated by sufficient linear algebra rank for matrix inversion was adopted uncritically in early gappy POD and DEIM work \cite{everson1995karhunen, chaturantabut2009discrete, drmac2016qdeim}, but might be the worst possible choice from the statistical point of view.

Prior work suggested mitigating the error with \emph{over-sampling} \cite{peherstorfer2020stability}, but our analysis of risk curves suggests that \emph{under-sampling} can also be a viable strategy. For the pseudoinverse reconstruction of the SST dataset with $r=100$ modes, $p\approx 80$ sensors yield a local minimum in error, and might be preferable if sensor budget is limited. More generally, the DNA framework can predict quantitatively the risk curves specific to a given dataset and help guide reconstruction design.

While in this paper we limited the reconstruction to linear methods, more complex nonlinear methods can be used, such as shallow recurrent decoders \cite{williams2024sensing, tomasetto2025reduced} and other neural networks. Machine Learning literature has demonstrated that double (or more generally multiple) descent is not unique to linear models and can easily arise in nonlinear models without proper regularization \cite{dascoli2020triple, transtrum2025generalized}.

\subsection{DEIM and time integration}
Our analysis of the DEIM algorithm revealed three parameters $r,q,p$ that do not need to be rigidly linked but can be selected independently. The central goal of constructing a ROM is to get a low error with minimal resources such as modes and sensors. Separation of resource accounting splits the question of ``how faithful are the ROM dynamics to the high-fidelity model?'' from ``how faithful is the sparse sensed nonlinear term to the fully sensed one?''. First, design of a good DEIM model needs to select an appropriate $r$ to represent the qualitative dynamics, such as the breather oscillation in the NSE system. Second, design needs to select $q$ such that the nonlinear term reconstruction projects well onto the dynamics of $r$ state modes. Third, in order to reconstruct the nonlinear term reliably, the double descent instability needs to be avoided by either a tuned regularization or oversampling $p>r$. Unlike the static sparse sensing case, undersampling $p<r$ is not going to work since it has a higher local error and thus the state error compounds faster.

Understanding the error compounding in ROMs is an important avenue for further work. In classical analysis of Initial Value Problems for ODEs, for a Lipshitz-continuous right hand side, the Lipshitz constant bounds how fast the solutions starting from nearby initial conditions can diverge from each other. By analogy, if the right hand side of a ROM has an upper-bounded error, how fast does it compound for the ROM to diverge from the high-fidelity model?

% how faithful are the dynamics vs how faithful is the nonlinear term
% can separate those discussions

% ROM question - how to get low error with minimal resources, as adding more sensors and modes relaxes the problem

% want enough r to resolve the dynamical system of interest - for NSE it is an oscillation. then want enough q so that the nonlinear modes project well enough onto the linear modes. then want either regularization or p>q to avoid the instability. unlike the static sparse sampling, undersampling p<q is not going to work because the error would compound too fast

% regularization is definitely useful

% in conventional numerical ODE solver analysis, for Lipshitz-continuous RHS, the Lipshitz constant bounds how fast can solutions starting form nearby initial conditions can diverge from each other
% analogy with Lipshitz constant
% what if the RHS is known within a relative error?

% POD gives three pieces of information
% 2018 only used singular vectors, we also use singular values
% reconstruction design - if you know what to expect for noise and contamination, need calibration
% primary study case was when noise and contamination are both present
% in general, the more is known
% if can't do regularization, can do under- or over-sampling. p=r is the worst case scenario

% maybe can do nonlinear, nonlinear double descent is known (can be even triple). depends on what is the 

% \paragraph{Time stepping and filtering}
% more involved time stepping

\subsection{Sensor set design}
For linear reconstructions as considered here, the reconstruction matrix $\mathbf{A}$ can be easily updated for any sensor set, and thus can be turned into a convenient Optimal Experimental Design objective to optimize a sensor set \cite{manohar2018data, chaloner1995bayesian, ryan2016review, alexanderian2021optimal, rainforth2024modern}. In this paper we introduced two testable competing hypotheses of which parts of the sensor set are to ``blame'' for instability. Depending on how sensors are placed, both 1-point and collective failures can be observed. Just like the placement landscape can in principle be decomposed into 1-sensor, 2-sensor, 3-sensor, and higher order interactions \cite{klishin2023data}, the collective failure hypothesis can be subdivided and investigated further. Perhaps a sensor placement algorithm that penalizes higher-order interactions would be able to avoid the collective failure for longer, yet no algorithm would be able to select more than $r$ orthogonal vectors in the $\mathbb{R}^r$ space.

While in a lot of prior work on sparse sensing and empirical interpolation \cite{willcox2006unsteady, yildirim2009efficient, chaturantabut2009discrete, drmac2016qdeim, manohar2018data} \emph{sensor placement} was posed as the central problem addressed by a variety of algorithms, our work suggests that \emph{reconstruction design through regularization} is a more significant factor for controlling the error. In other words, a critical task is not to find the \emph{best} sensor set, but merely one that is \emph{good enough}: for example, the 2-point method used here and the QR pivoting method \cite{drmac2016qdeim, manohar2018data} approximately optimize nearly the same objective and thus result in similar error metrics \cite{klishin2023data}. Relaxing the need to optimize the sensor set allows for much easier incorporation of non-data-driven objectives, such as sensor costs and constraints \cite{karnik2024optimal}, or deliberate redundancy to provide backups in case of sensor failure.

% who to blame - 1-pt or collective
% can extend this to more specific combinations
% collective sensor design beyond 2-pt - might be possible with the determinant expansion but pointless, using regularization or undersampling is better. can model with a singular value spectrum
% incorporate other non-data driven costs and constraints, e.g. karnik or deliberate sensor redundancy

\subsection{Uncertainty quantification}
The \emph{Bayesian} nature of the reconstruction implies not only a regularization on the inference of the mean (Maximum A Posteriori \cite{Kakasenko2026BridgingTG}), but a full probability distribution around the most likely state. This probability distribution promises a statistical guarantee of error distribution demanded by safety-critical engineering applications. Our earlier work provided a measure of uncertainty quantification through \emph{uncertainty heatmaps} given by the diagonal part of the error covariance, but found the error to be significantly under-estimated \cite{klishin2023data}. While that work only considered the error driven by noise (our $\mathbf{B}_3$ term), here we extended the calculation to four terms of Eqn.~\ref{eqn:K}, which can be straightforwardly converted into an updated pixel-wise heatmap. Since the covariance terms are positive-definite, the elements of a more comprehensive uncertainty heatmap are definitely larger. It remains to be seen in future work if the updated uncertainty is appropriately calibrated and applicable to the range of reconstruction scenarios.

% can do UQ mode-wise instead of point-wise, study different contributors in the formula
% trace vs diagonal entries

% \begin{figure*}[htbp]
%   \centering
%   \label{fig:a}\includegraphics{lexample_fig1}
%   \caption{Example figure using external image files.}
%   \label{fig:testfig}
% \end{figure*}

\section*{Acknowledgments}
The authors would like to thank J.~Bechhoefer, Y.~Bhangale, L.~Fujioka, N.~Karnik, J.~Kent-Dobias, and S.~E.~Otto for helpful discussions and L.~D. Lederer for administrative support. This work uses Scientific Color Maps for visualization \cite{crameri2023color} and the \texttt{Signac} framework for computational data management \cite{adorf2018simple}. Generative Artificial Intelligence was not used for the preparation of text, code, or figures. The authors acknowledge support from the National Science Foundation AI Institute in Dynamic Systems (grant number 2112085) and MATH DT program (grant numbers 2529361/2529362).

% \clearpage
\bibliography{references}

\end{document}